\documentclass[lettersize,journal]{IEEEtran}
\usepackage{amsmath,amsfonts}
\usepackage{array}
\usepackage[caption=false,font=normalsize,labelfont=sf,textfont=sf]{subfig}
\usepackage{textcomp}
\usepackage{stfloats}
\usepackage{url}
\usepackage{verbatim}
\usepackage{graphicx}
\usepackage{cite}
\usepackage{times}
\usepackage{epsfig}
\usepackage{graphicx}
\usepackage{amsmath}
\usepackage{amssymb}
\usepackage{xcolor}
\usepackage{booktabs, makecell, tabularx}
\usepackage{multirow}
\usepackage{dsfont}
\usepackage{enumitem}
\usepackage{algorithm}
\usepackage{algpseudocode}
\usepackage{xspace}
\usepackage{caption,cite}

\usepackage{pifont}%

\definecolor{mygray}{gray}{.93}
\definecolor{mygray1}{gray}{.99}

\def\eg{\textit{e.g.}}
\def\etc{\textit{etc}}

\definecolor{urlcolor}{RGB}{255,105,180}
\definecolor{citecolor}{RGB}{0, 80, 200}
\definecolor{linkcolor}{HTML}{ED1C24}
\definecolor{highlightcolor}{HTML}{ABCDEF}

\usepackage[pagebackref=false,breaklinks=true,letterpaper=true,colorlinks,bookmarks=true]{hyperref}

\newlength\secmargin
\newlength\subsecmargin
\newlength\paramargin
\newlength\figmargin
\newlength\eqmargin
\usepackage[capitalize]{cleveref}
\crefname{section}{Sec.}{Secs.}
\Crefname{section}{Section}{Sections}
\Crefname{table}{Table}{Tables}
\crefname{table}{Tab.}{Tabs.}
\crefname{algorithm}{Algo.}{Algos.}

\usepackage{makecell}
\usepackage{colortbl}
\usepackage{booktabs}

\usepackage{bm}

\providecommand{\impath}[1]{}
\providecommand{\impatha}[1]{}
\providecommand{\impathb}[1]{}
\providecommand{\impathc}[1]{}
\providecommand{\impathd}[1]{}
\providecommand{\impathe}[1]{}

\newcommand{\figteaser}{
    \centering

    \makebox[0.115\textwidth]{[Frame-wise Image Editing Result] A dog is walking on the ground, \colorbox{highlightcolor}{Van Gogh style}} \\
    \includegraphics[width=0.115\textwidth]{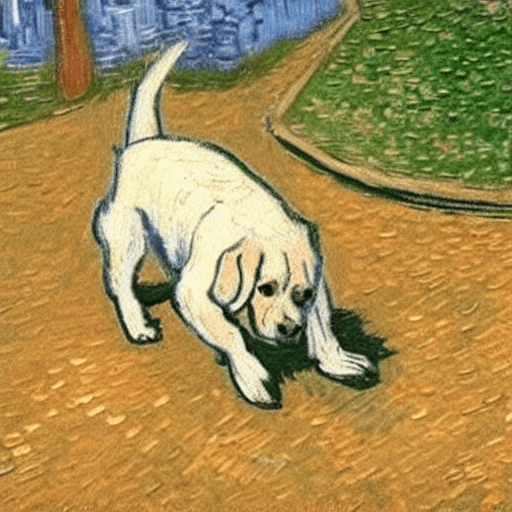}
    \includegraphics[width=0.115\textwidth]{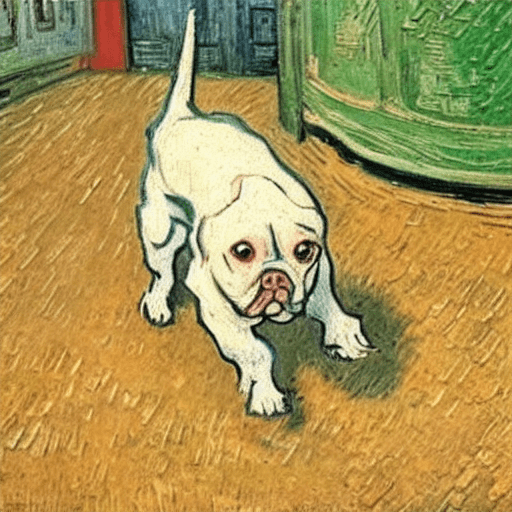}
    \includegraphics[width=0.115\textwidth]{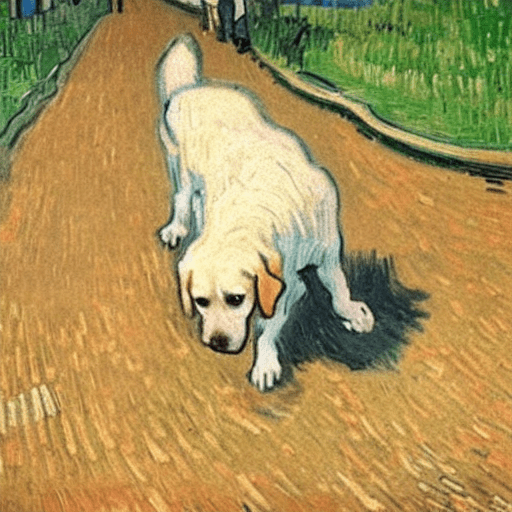}
    \includegraphics[width=0.115\textwidth]{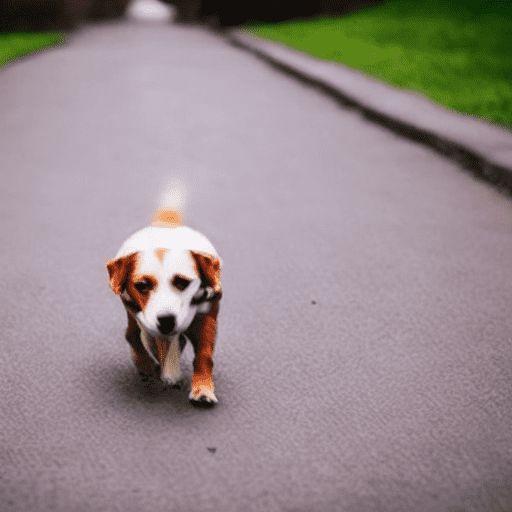}
    \includegraphics[width=0.115\textwidth]{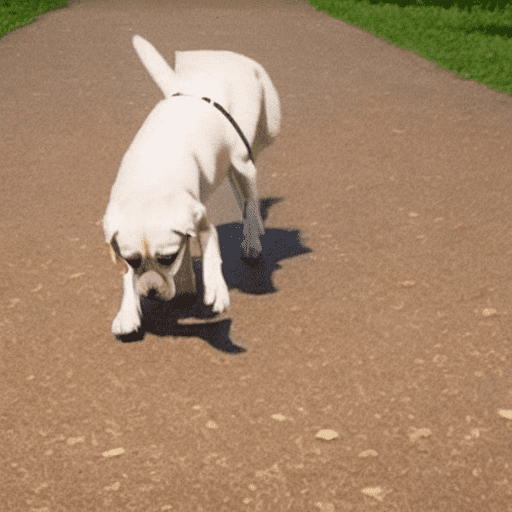}
    \includegraphics[width=0.115\textwidth]{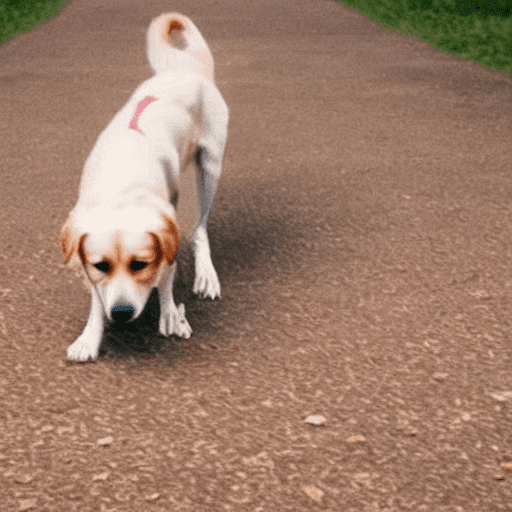}
    \includegraphics[width=0.115\textwidth]{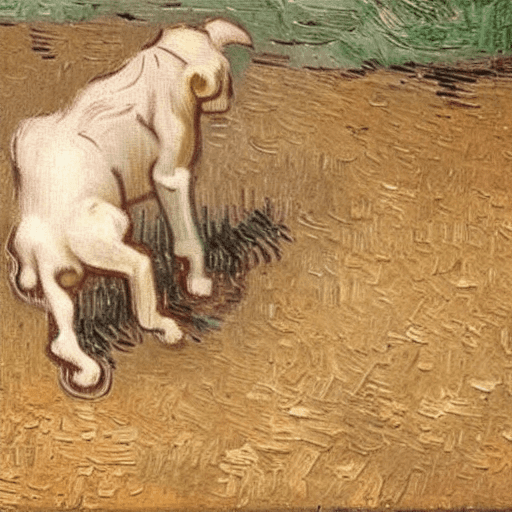}
    \includegraphics[width=0.115\textwidth]{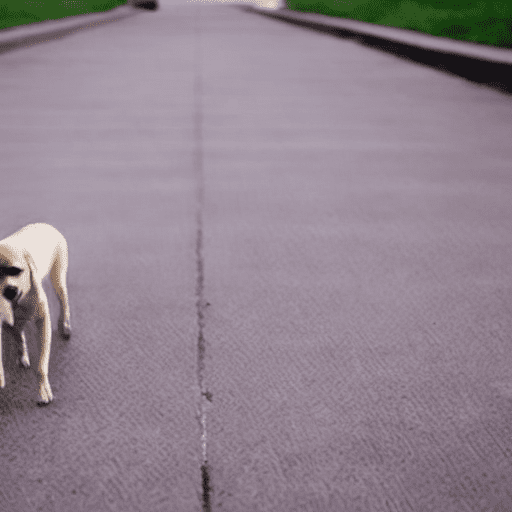}
    
    \makebox[0.115\textwidth]{[Input Video] A dog is walking on the ground.}\\
    \includegraphics[width=0.115\textwidth]{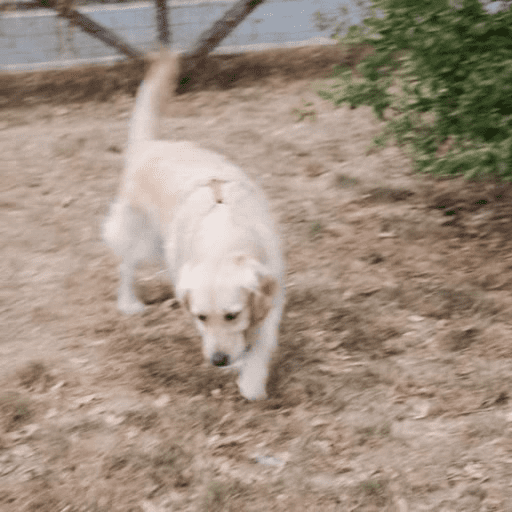}
    \includegraphics[width=0.115\textwidth]{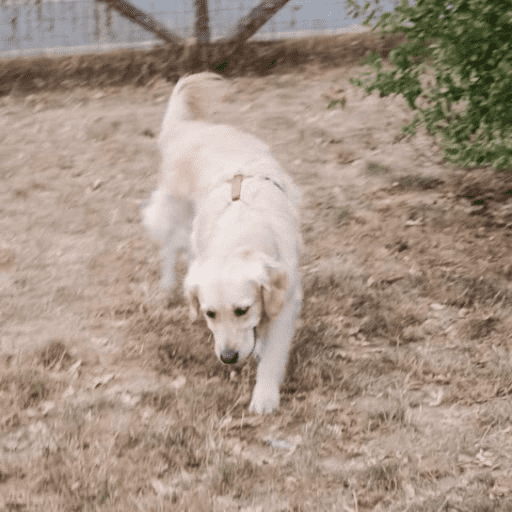}
    \includegraphics[width=0.115\textwidth]{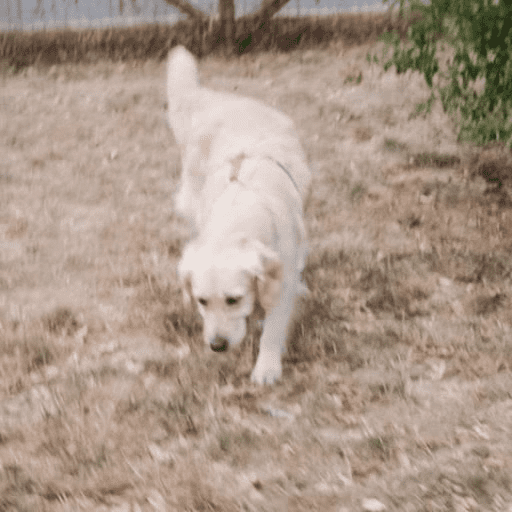}
    \includegraphics[width=0.115\textwidth]{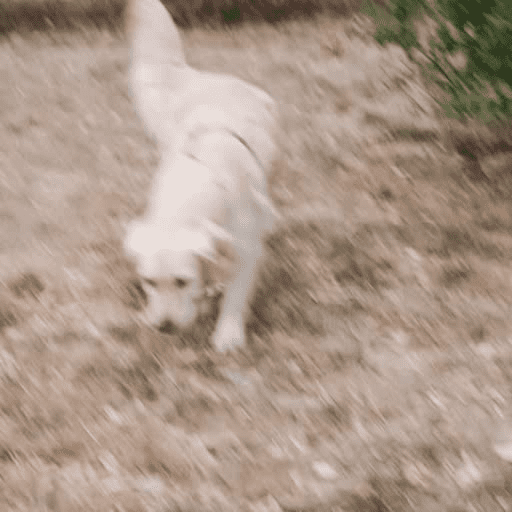}
    \includegraphics[width=0.115\textwidth]{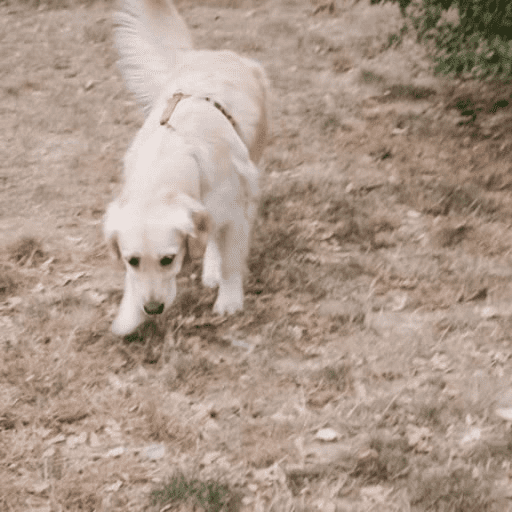}
    \includegraphics[width=0.115\textwidth]{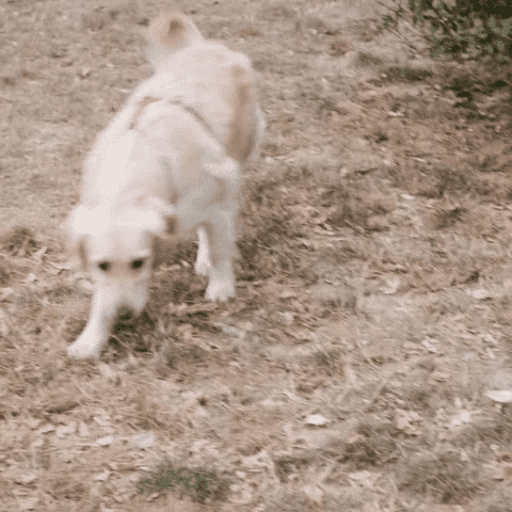}
    \includegraphics[width=0.115\textwidth]{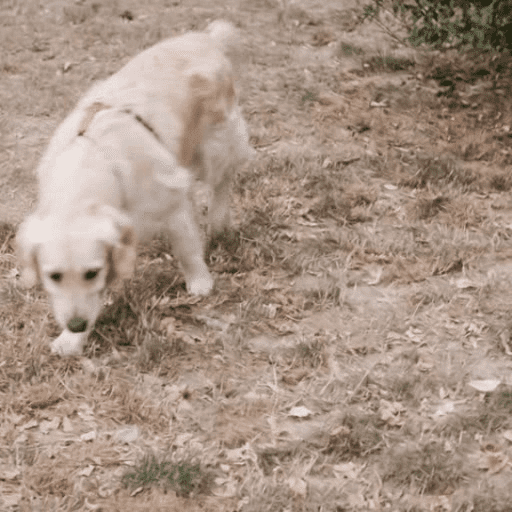}
    \includegraphics[width=0.115\textwidth]{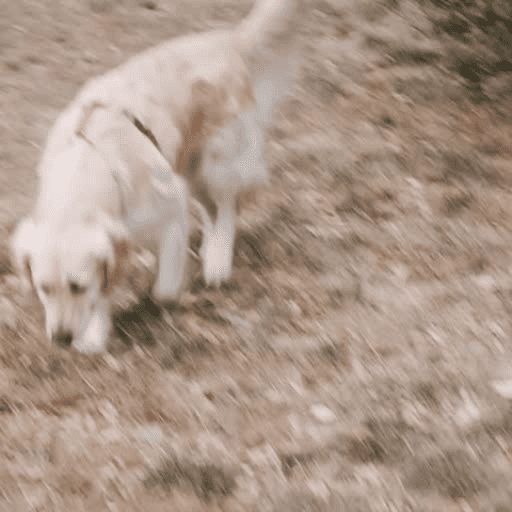}

    \makebox[0.115\textwidth]{[Our Editing Result] A dog is walking on the ground, \colorbox{highlightcolor}{Van Gogh style}} \\
    \includegraphics[width=0.115\textwidth]{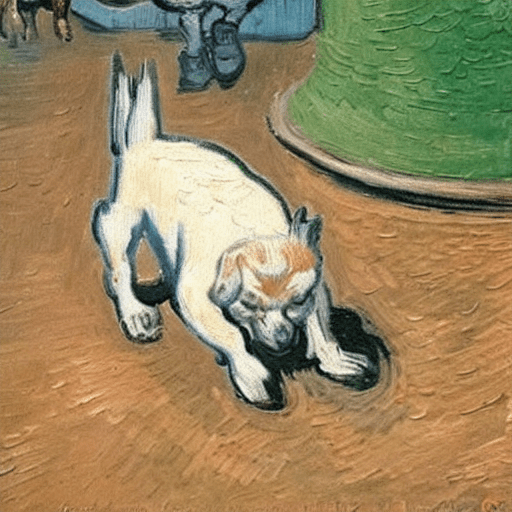}
    \includegraphics[width=0.115\textwidth]{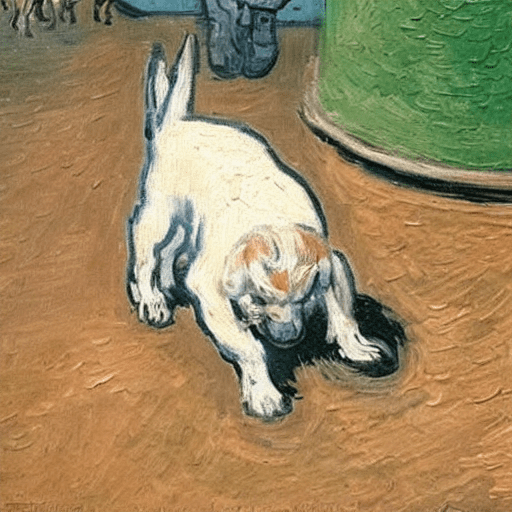}
    \includegraphics[width=0.115\textwidth]{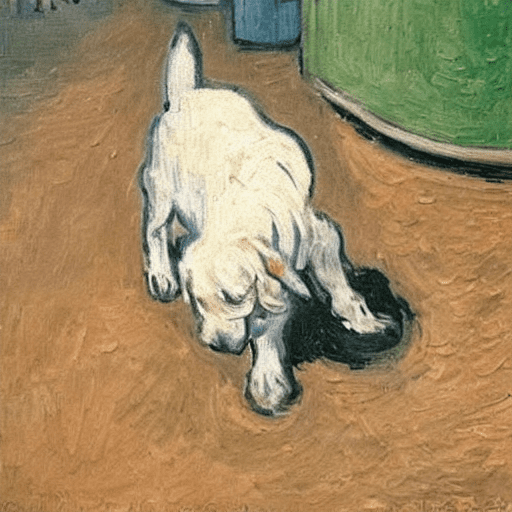}
    \includegraphics[width=0.115\textwidth]{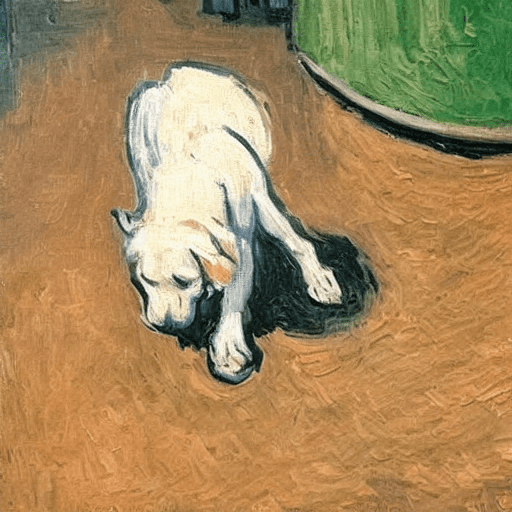}
    \includegraphics[width=0.115\textwidth]{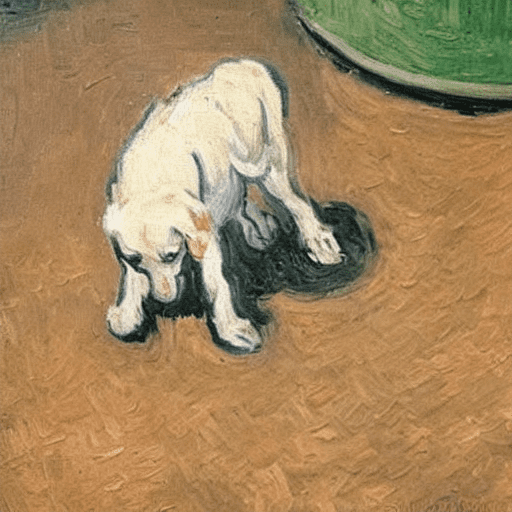}
    \includegraphics[width=0.115\textwidth]{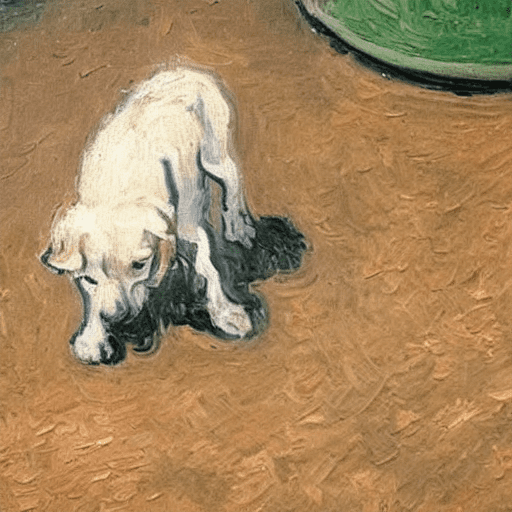}
    \includegraphics[width=0.115\textwidth]{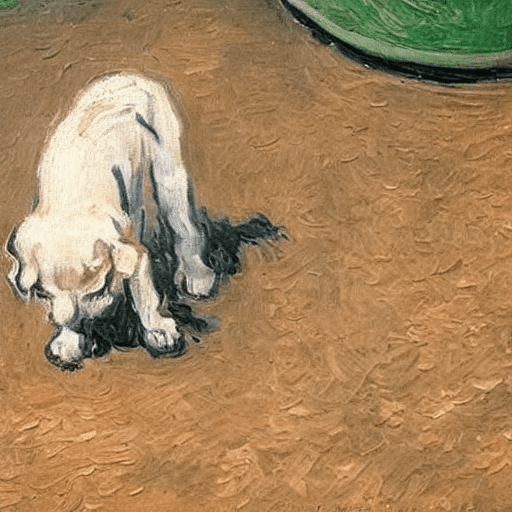}
    \includegraphics[width=0.115\textwidth]{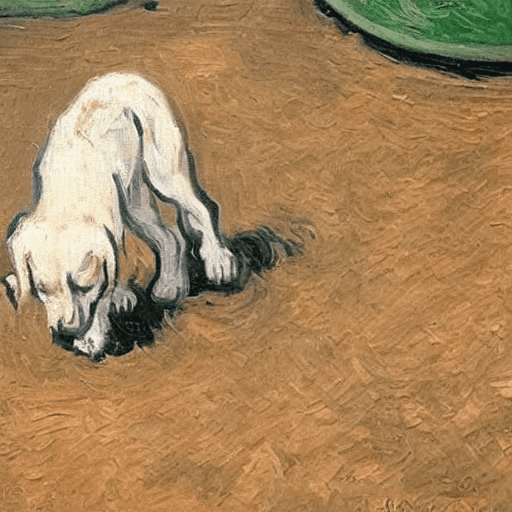}
    
    \captionof{figure}{Video editing using 
    an 
    \textbf{off-the-shelf} image diffusion model. The straightforward frame-wise image editing results in severe flickering effects (first row). 
    In contrast, our method achieves temporally-consistent editing results (third row). 
    }
    \label{fig:teaser}

    \vspace{0.8em}
}

\newcommand{\figmethod}{
    \begin{figure*}[t]
      \centerline{\includegraphics[width=\textwidth]{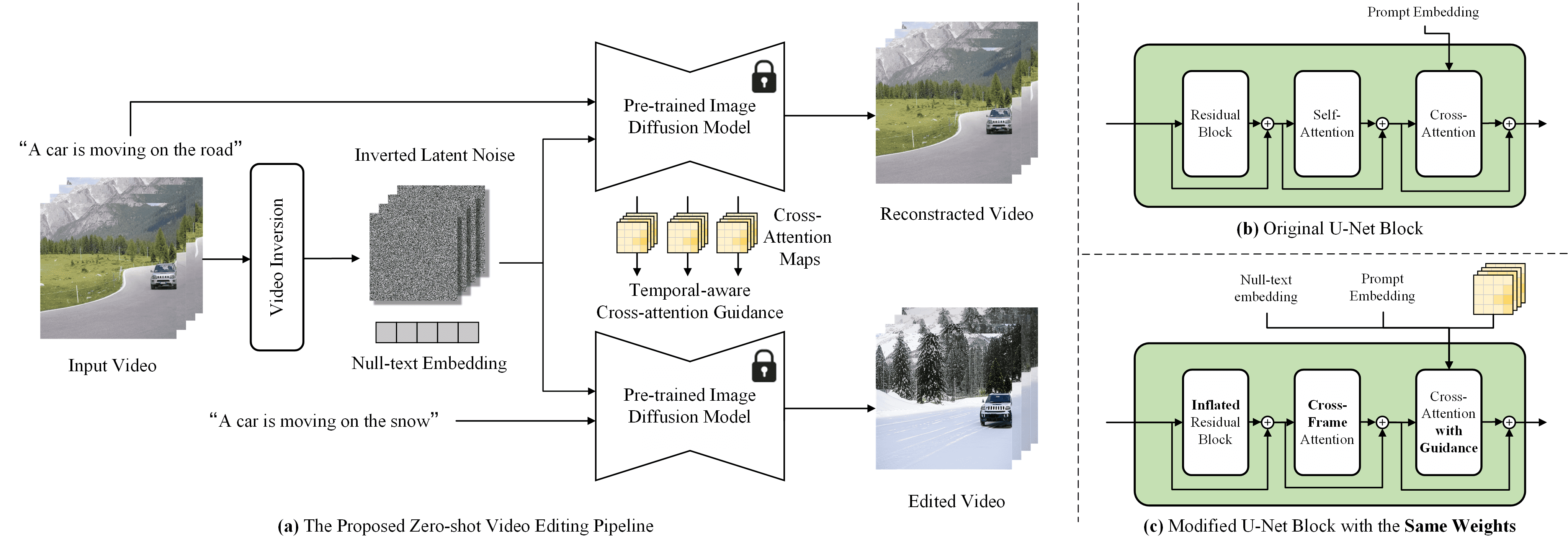}}
      \caption{Illustration of the proposed vid2vid-zero for zero-shot video editing. (a) Our framework first inverses the input video to obtain the latent noise and null-text embedding, then use the inversion results to generate the edited video, under cross-attention guidance. Temporal modeling is achieved by replacing self-attention in the original U-Net block (b) with the cross-frame attention in (c) that shares the same model weights.}
      \label{fig:method}
    \end{figure*}
}

\newcommand{\figattntypes}{
    \begin{figure}[t]
      \centerline{\includegraphics[width=0.5\textwidth]{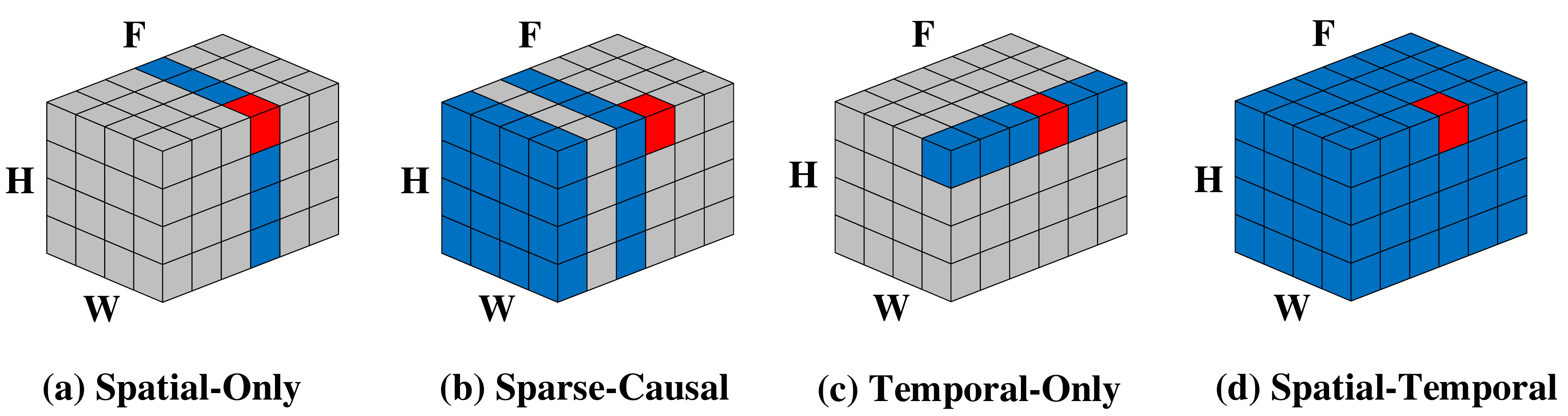}}
      \caption{Illustration of different attention mechanisms. The query and keys are highlighted in red and blue, respectively. $H$, $W$, and $F$ represent the height, width, and temporal dimension of the input video.}
      \label{fig:attn_type}
    \end{figure}
}

\newcommand{\figuser}{
    \begin{figure}[t]
      \centerline{\includegraphics[width=0.5\textwidth]{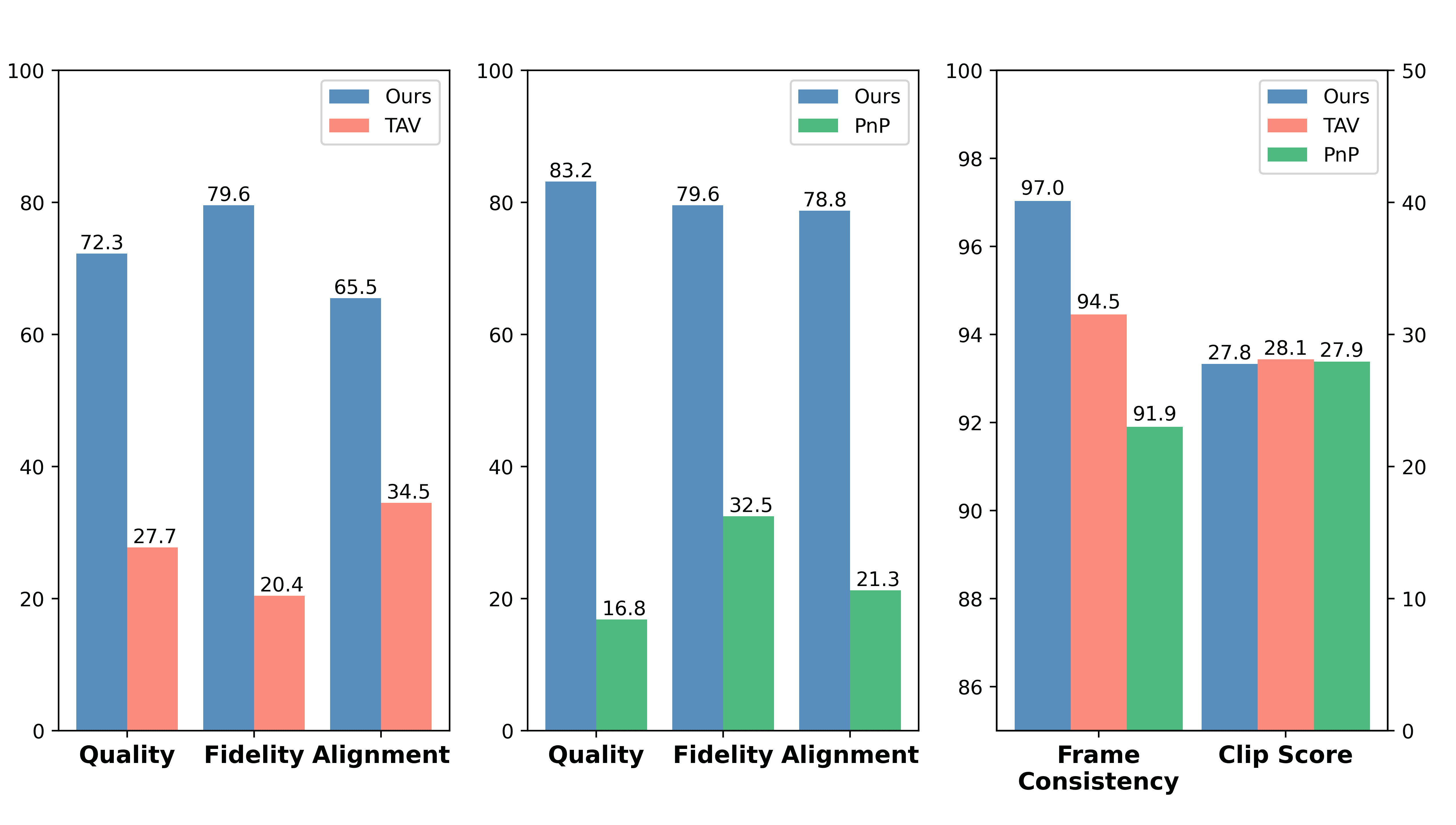}}
      \caption{Quantitative results. User preference studies compared to TAV and PnP are shown in the first and second columns, respectively. CLIP score and frame consistency are presented in the third column.}
      \label{fig:user}
    \end{figure}
}

\newcommand{\figresults}{
    \begin{figure*}[p]
    \centering

    \makebox[0.115\textwidth]{[Input Video] A man is running.} \\
    \includegraphics[width=0.115\textwidth]{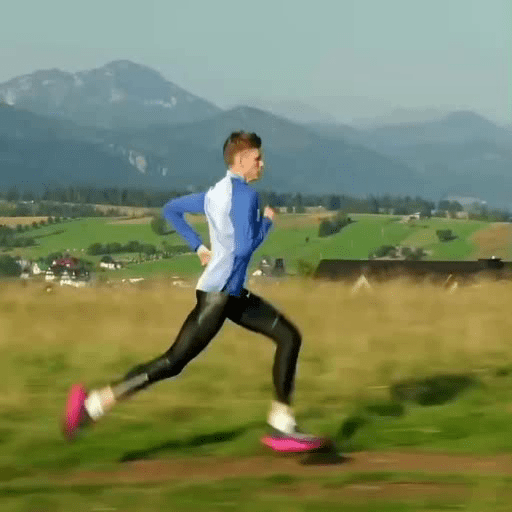}
    \includegraphics[width=0.115\textwidth]{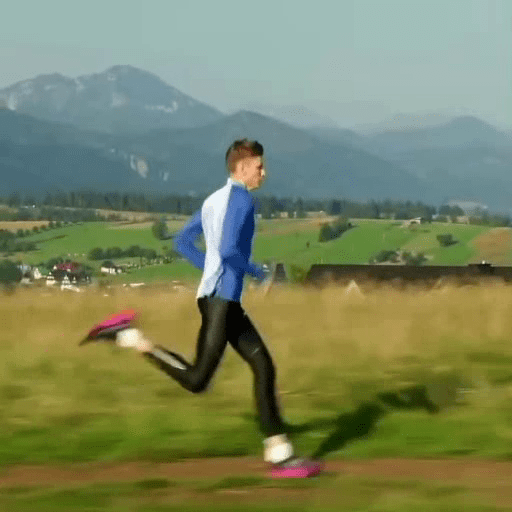}
    \includegraphics[width=0.115\textwidth]{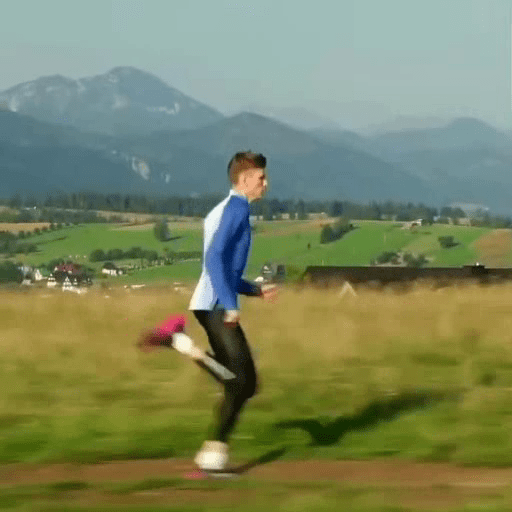}
    \includegraphics[width=0.115\textwidth]{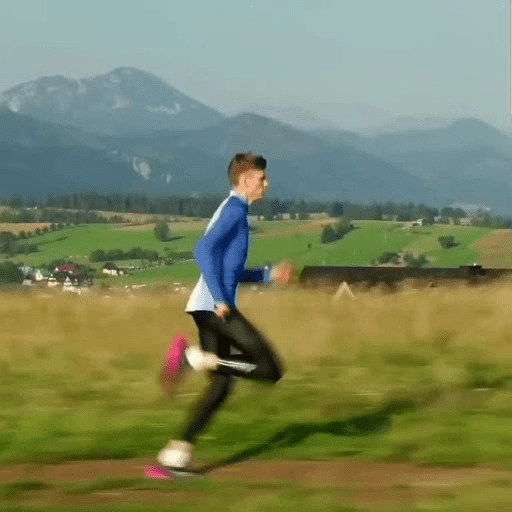}
    \includegraphics[width=0.115\textwidth]{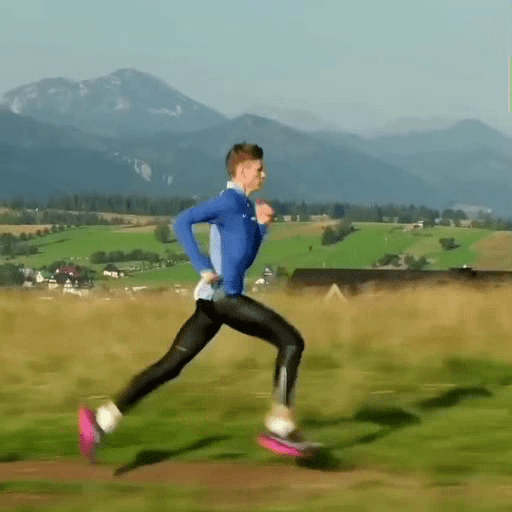}
    \includegraphics[width=0.115\textwidth]{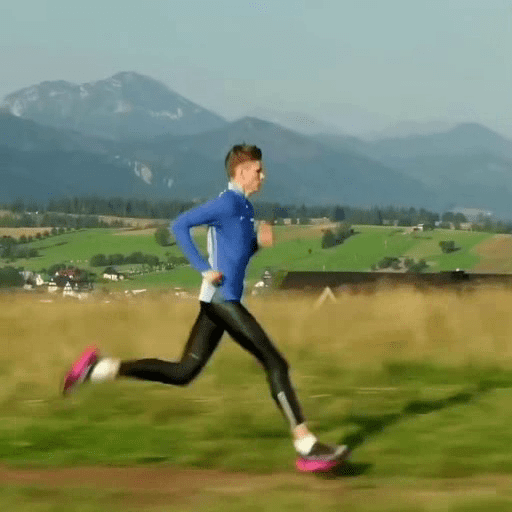}
    \includegraphics[width=0.115\textwidth]{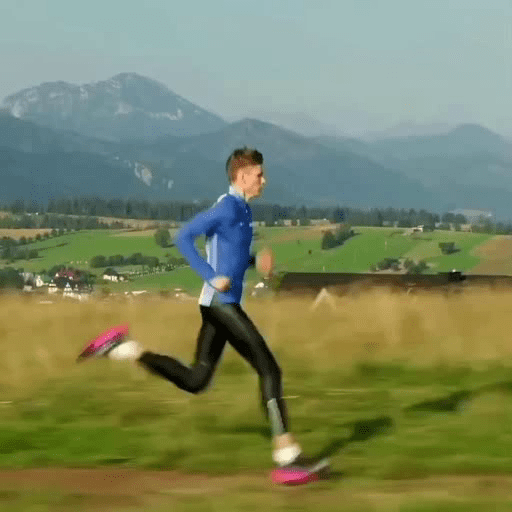}
    \includegraphics[width=0.115\textwidth]{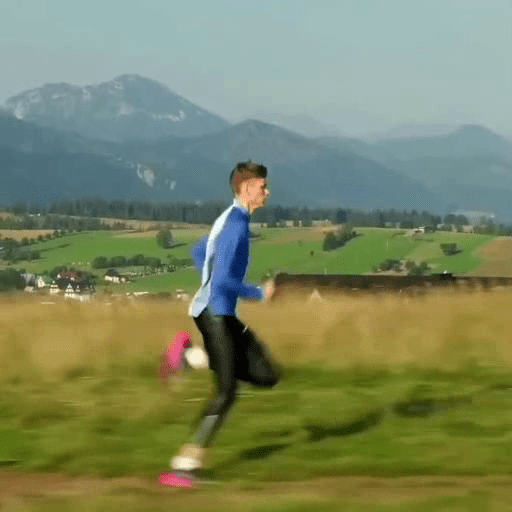}

    \makebox[0.115\textwidth]{A man is running, \colorbox{highlightcolor}{anime style.}} \\
    \includegraphics[width=0.115\textwidth]{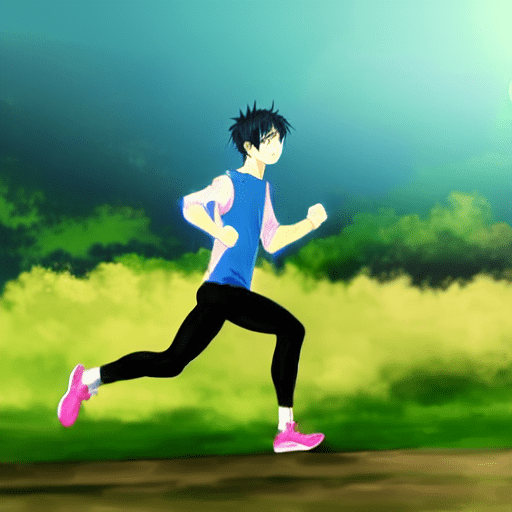}
    \includegraphics[width=0.115\textwidth]{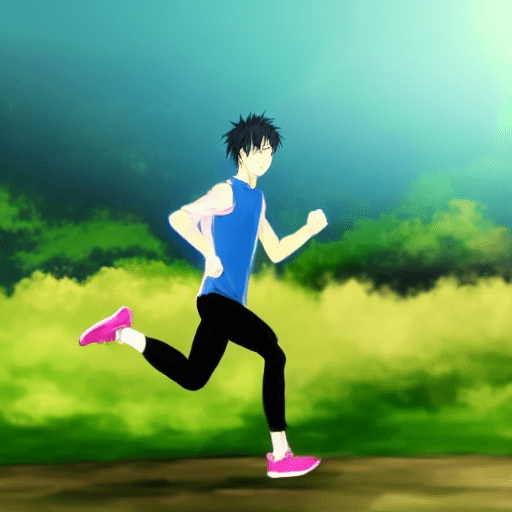}
    \includegraphics[width=0.115\textwidth]{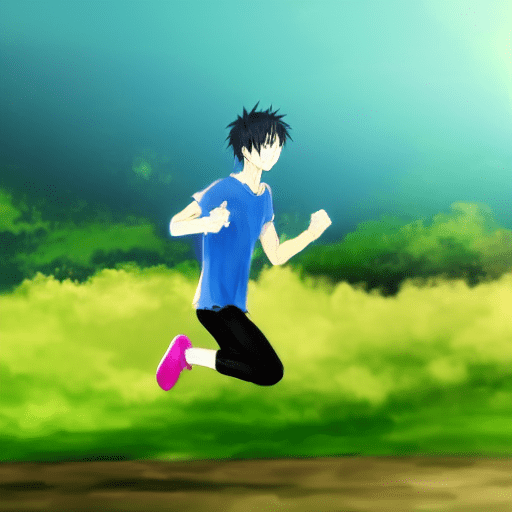}
    \includegraphics[width=0.115\textwidth]{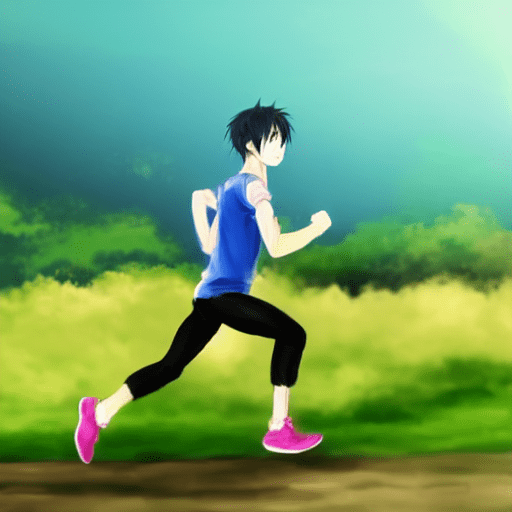}
    \includegraphics[width=0.115\textwidth]{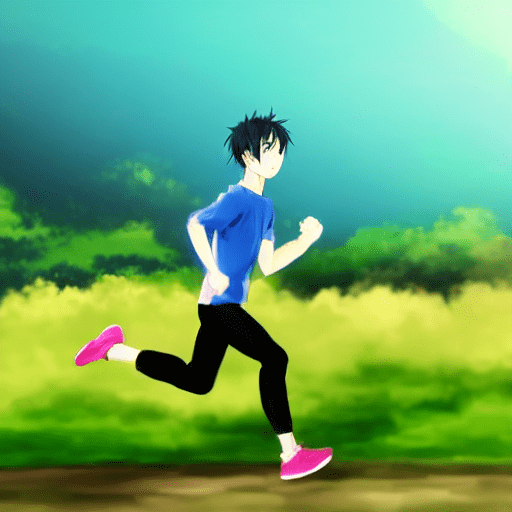}
    \includegraphics[width=0.115\textwidth]{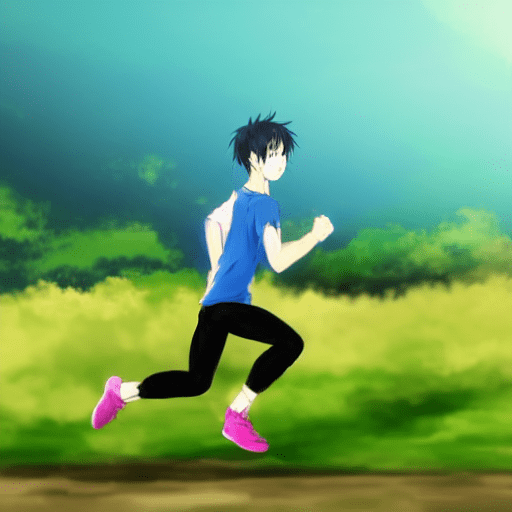}
    \includegraphics[width=0.115\textwidth]{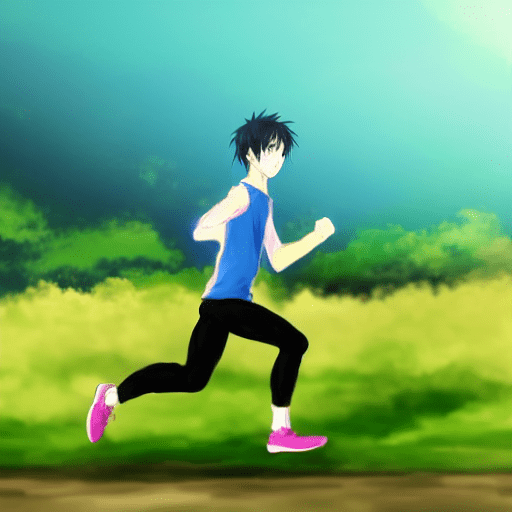}
    \includegraphics[width=0.115\textwidth]{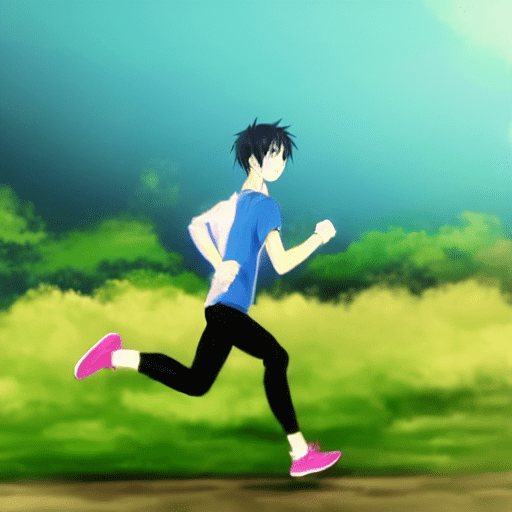}

    \makebox[0.115\textwidth]{\colorbox{highlightcolor}{An old man} is running \colorbox{highlightcolor}{on the beach.}} \\
    \includegraphics[width=0.115\textwidth]{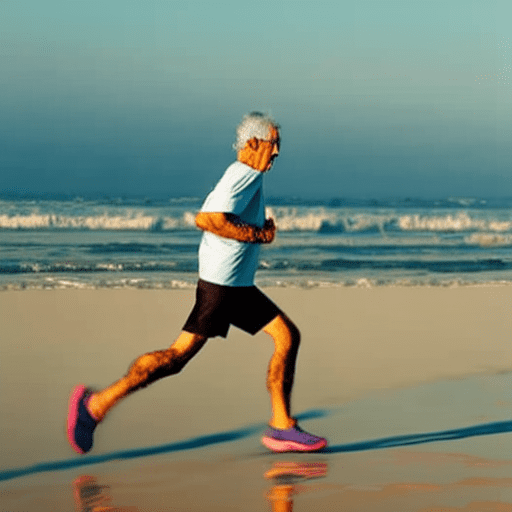}
    \includegraphics[width=0.115\textwidth]{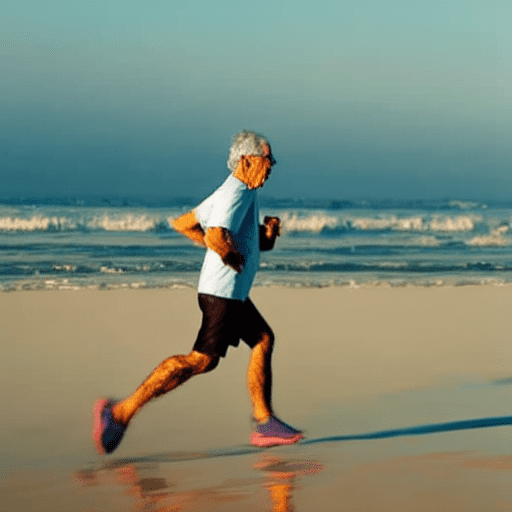}
    \includegraphics[width=0.115\textwidth]{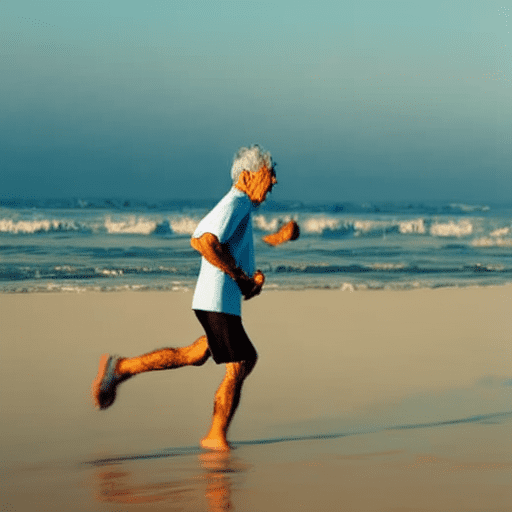}
    \includegraphics[width=0.115\textwidth]{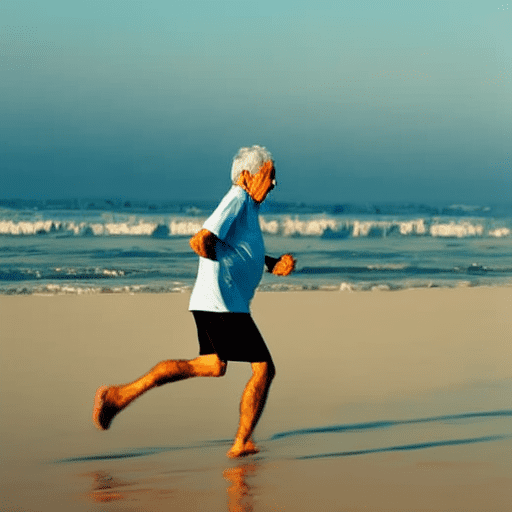}
    \includegraphics[width=0.115\textwidth]{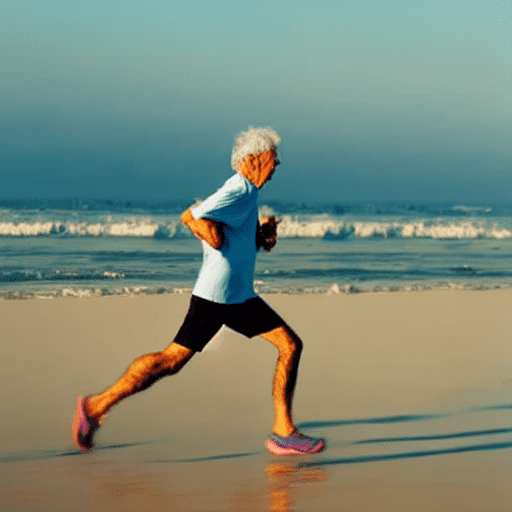}
    \includegraphics[width=0.115\textwidth]{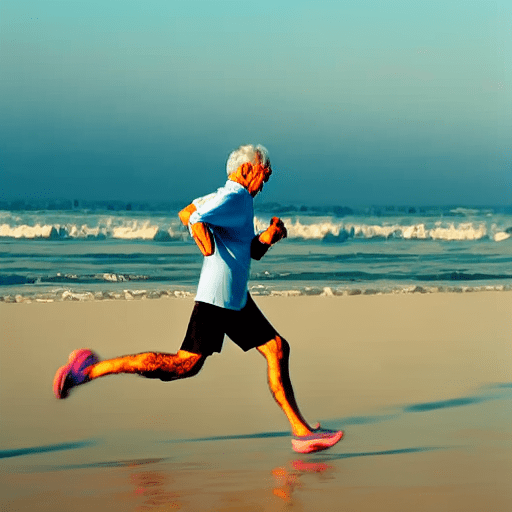}
    \includegraphics[width=0.115\textwidth]{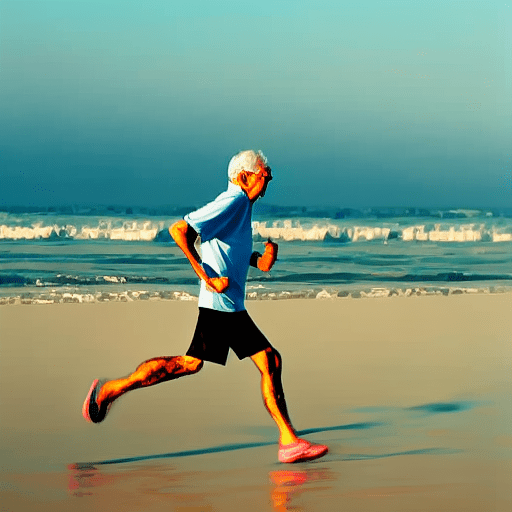}
    \includegraphics[width=0.115\textwidth]{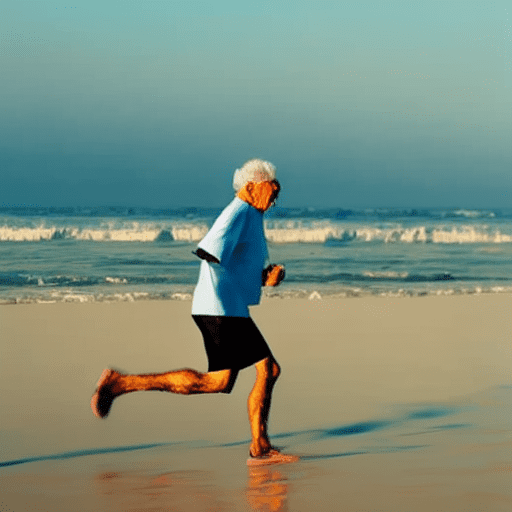}

    \makebox[0.115\textwidth]{\colorbox{highlightcolor}{Stephen Curry} is running \colorbox{highlightcolor}{in Time Square.}} \\
    \includegraphics[width=0.115\textwidth]{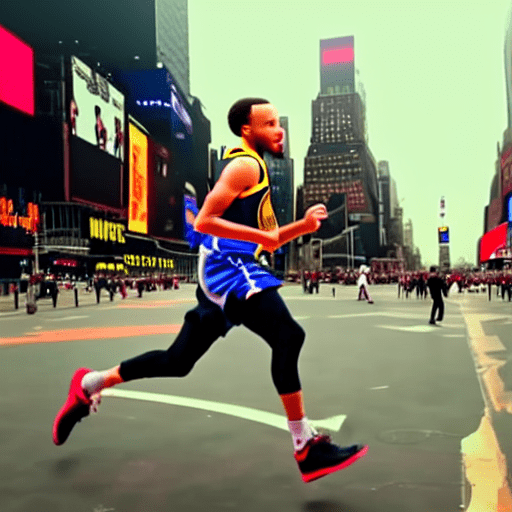}
    \includegraphics[width=0.115\textwidth]{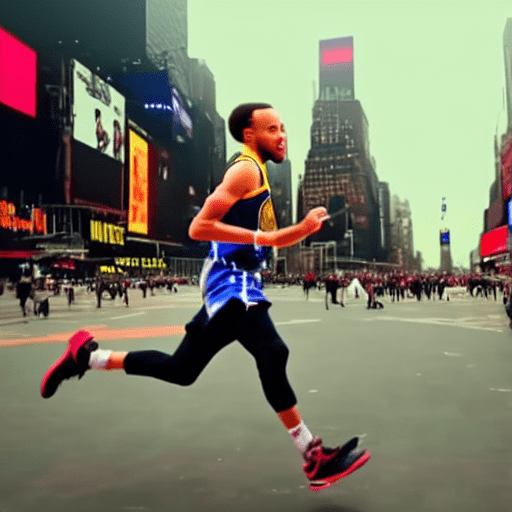}
    \includegraphics[width=0.115\textwidth]{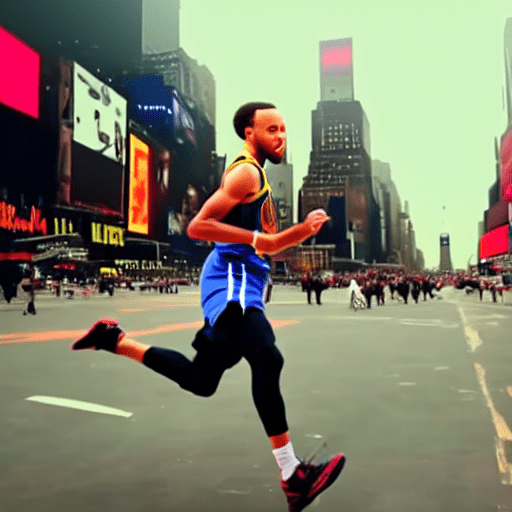}
    \includegraphics[width=0.115\textwidth]{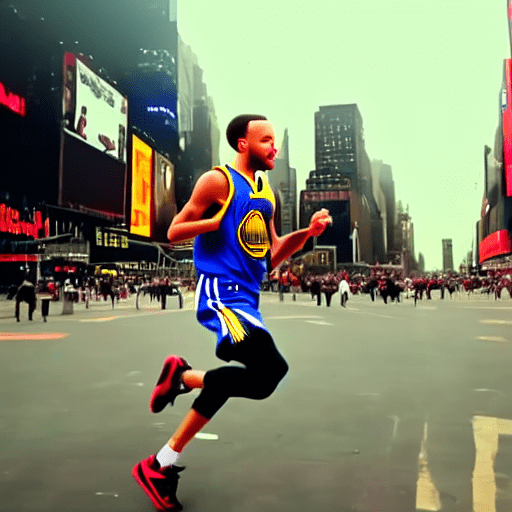}
    \includegraphics[width=0.115\textwidth]{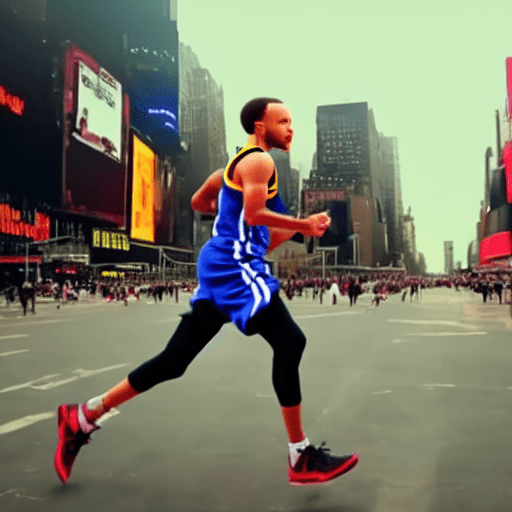}
    \includegraphics[width=0.115\textwidth]{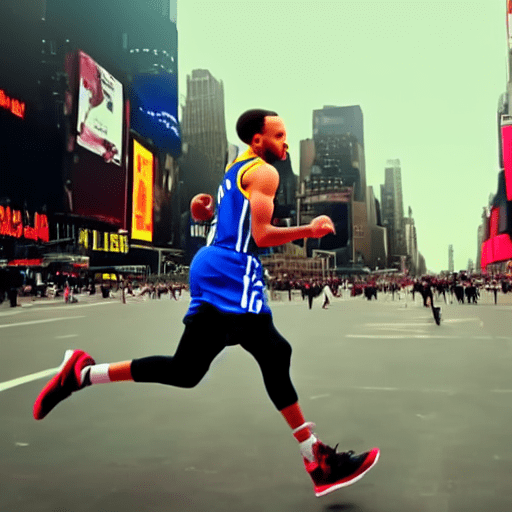}
    \includegraphics[width=0.115\textwidth]{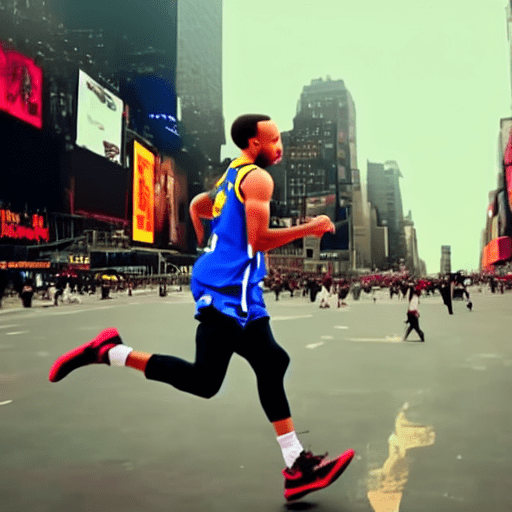}
    \includegraphics[width=0.115\textwidth]{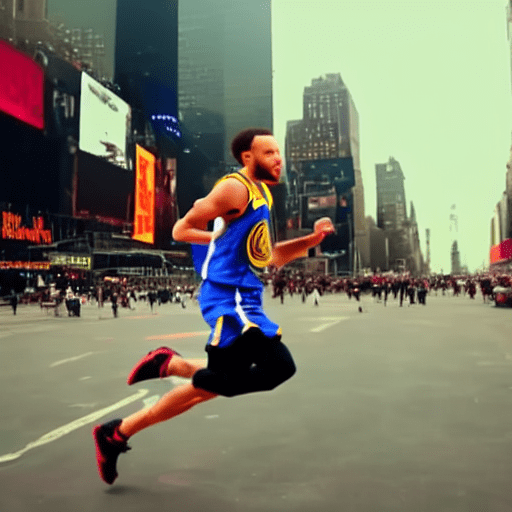}
    
    \makebox[0.115\textwidth]{} 

    \makebox[0.115\textwidth]{[Input Video] A car is moving on the road.} \\
    \includegraphics[width=0.115\textwidth]{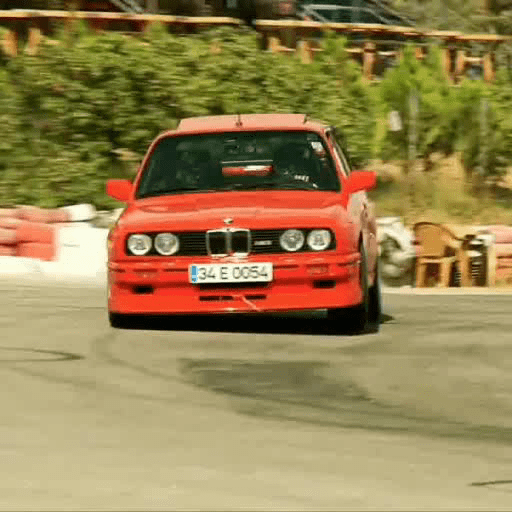}
    \includegraphics[width=0.115\textwidth]{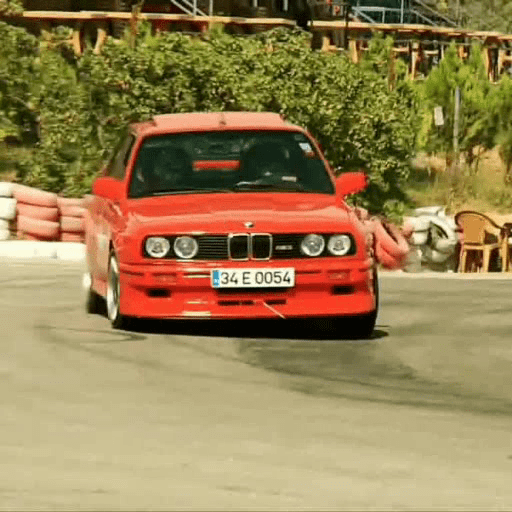}
    \includegraphics[width=0.115\textwidth]{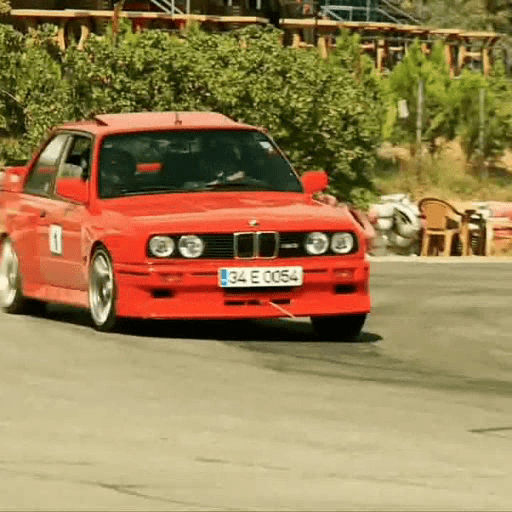}
    \includegraphics[width=0.115\textwidth]{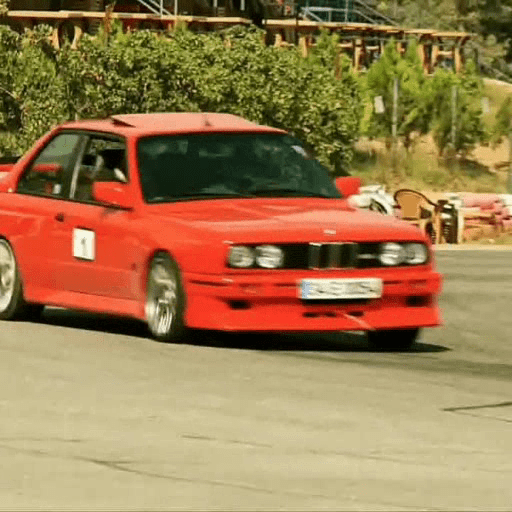}
    \includegraphics[width=0.115\textwidth]{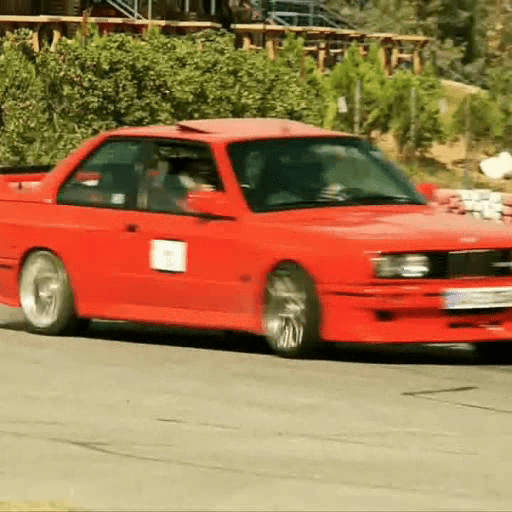}
    \includegraphics[width=0.115\textwidth]{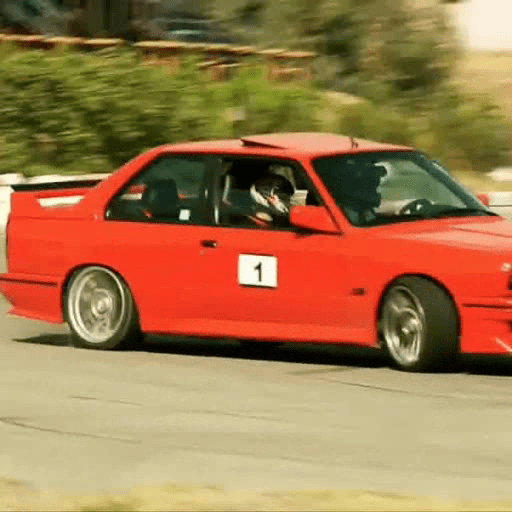}
    \includegraphics[width=0.115\textwidth]{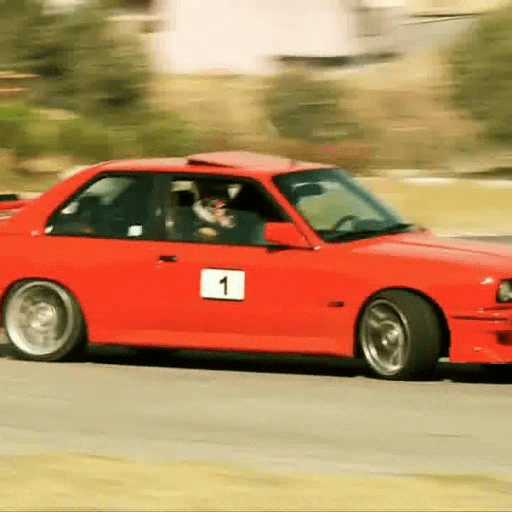}
    \includegraphics[width=0.115\textwidth]{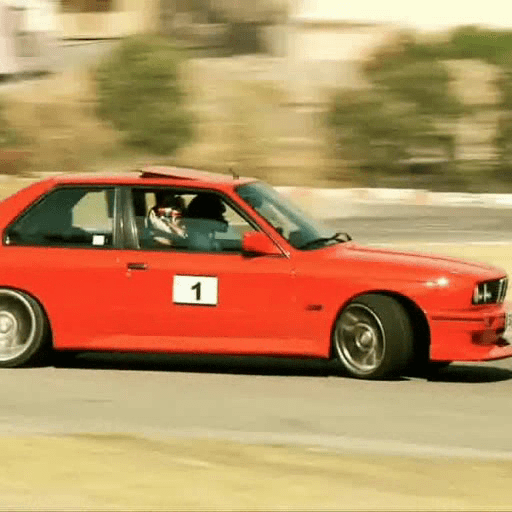}

    \makebox[0.115\textwidth]{A car is moving on the \colorbox{highlightcolor}{snow}.}\\
    \includegraphics[width=0.115\textwidth]{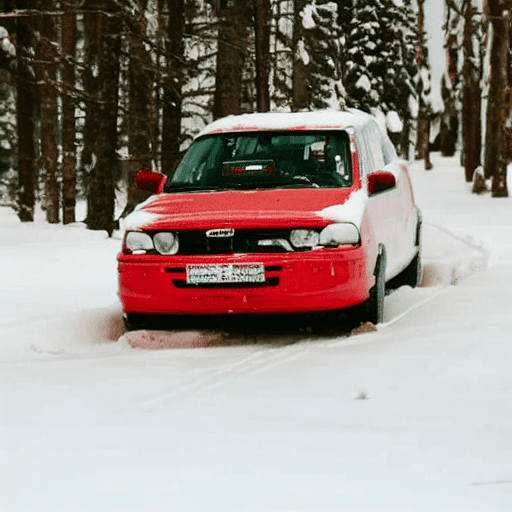}
    \includegraphics[width=0.115\textwidth]{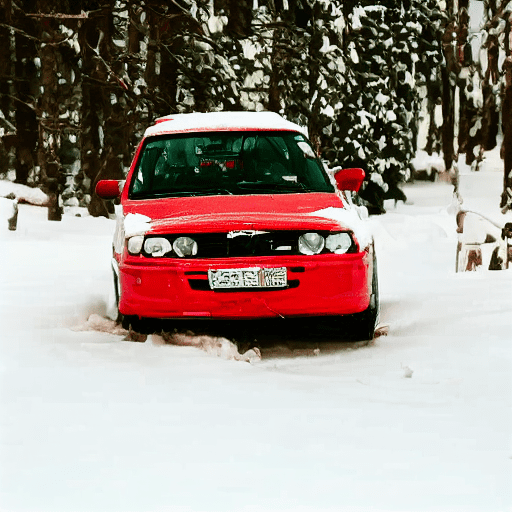}
    \includegraphics[width=0.115\textwidth]{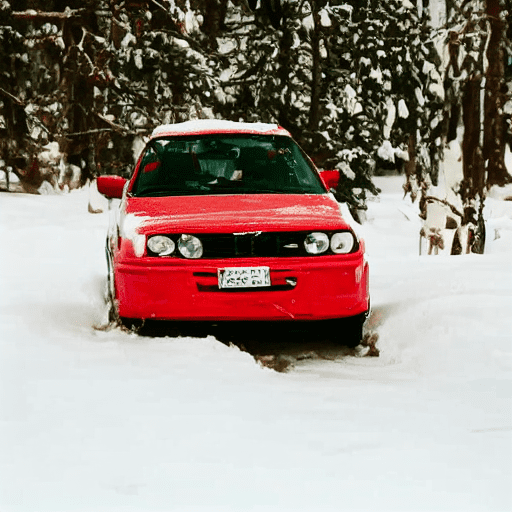}
    \includegraphics[width=0.115\textwidth]{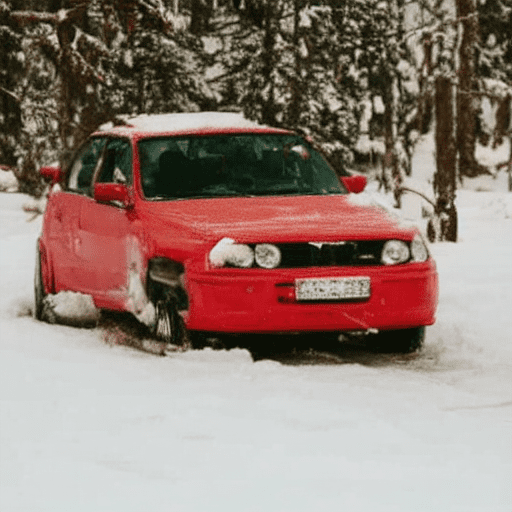}
    \includegraphics[width=0.115\textwidth]{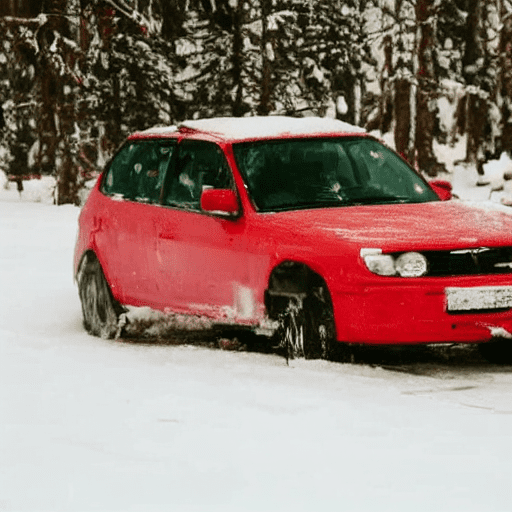}
    \includegraphics[width=0.115\textwidth]{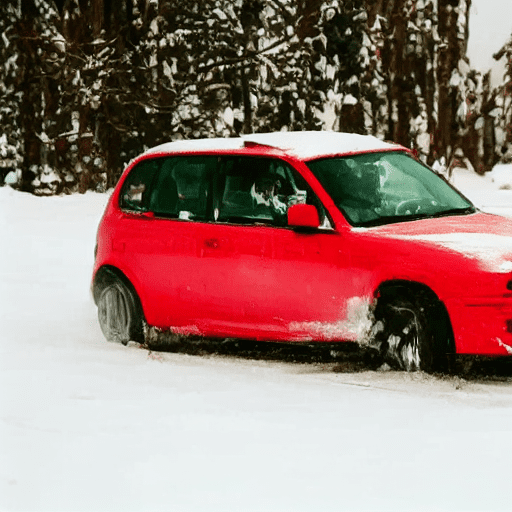}
    \includegraphics[width=0.115\textwidth]{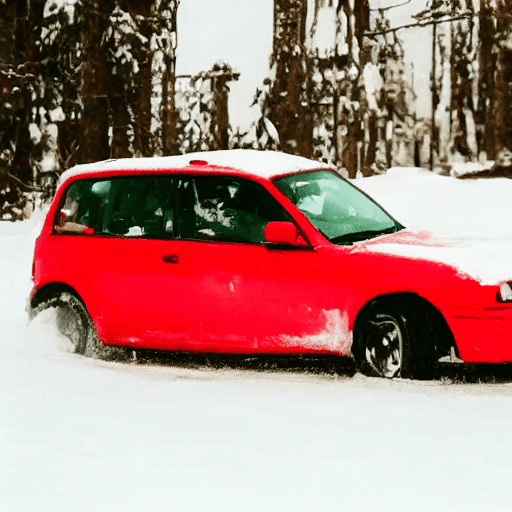}
    \includegraphics[width=0.115\textwidth]{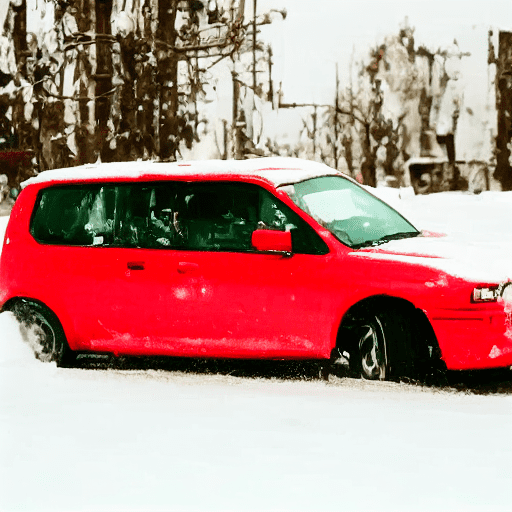}
    
    \makebox[0.115\textwidth]{A \colorbox{highlightcolor}{jeep car} is moving on the road.} \\
    \includegraphics[width=0.115\textwidth]{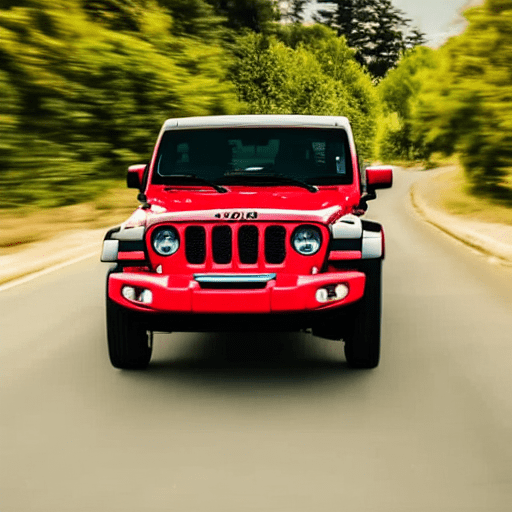}
    \includegraphics[width=0.115\textwidth]{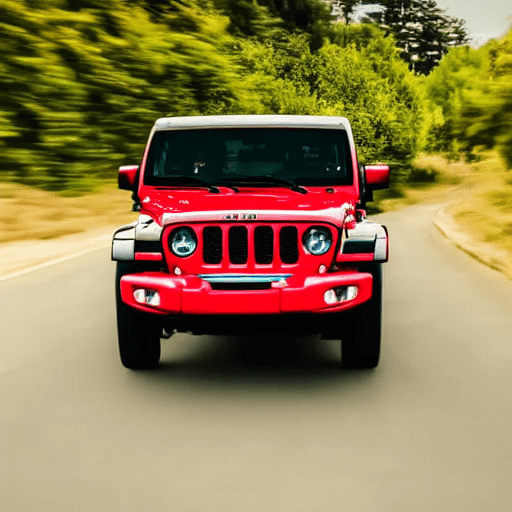}
    \includegraphics[width=0.115\textwidth]{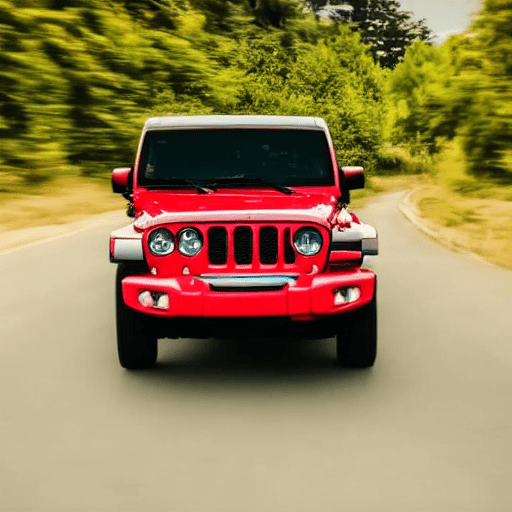}
    \includegraphics[width=0.115\textwidth]{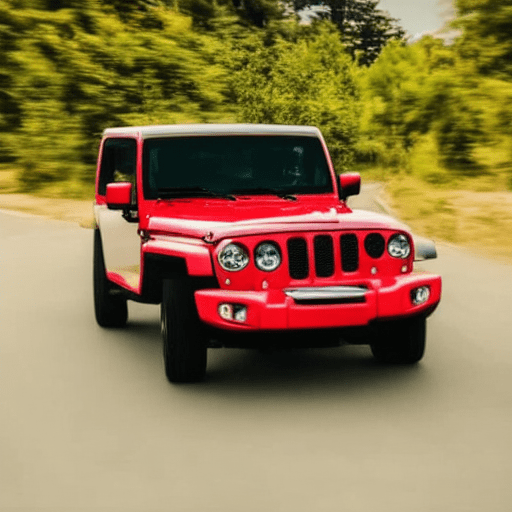}
    \includegraphics[width=0.115\textwidth]{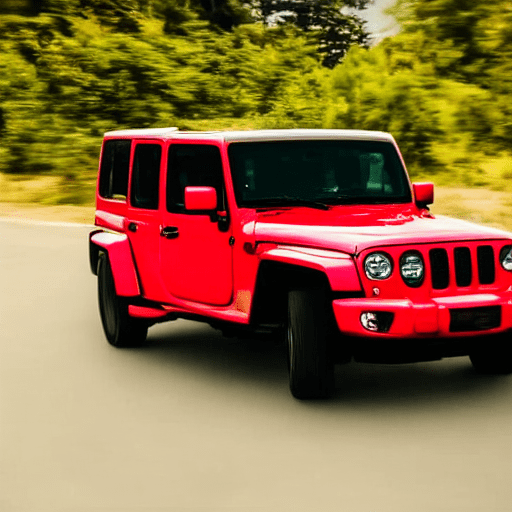}
    \includegraphics[width=0.115\textwidth]{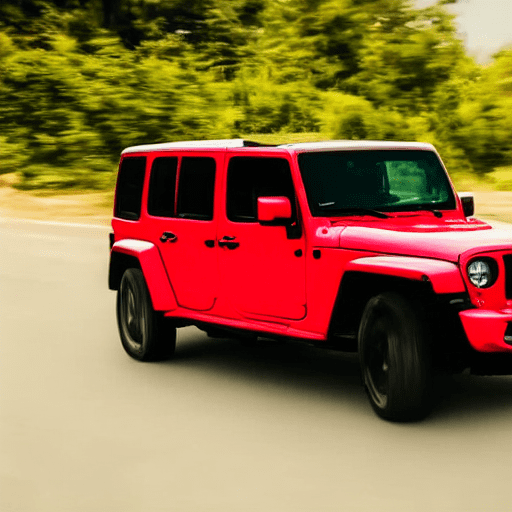}
    \includegraphics[width=0.115\textwidth]{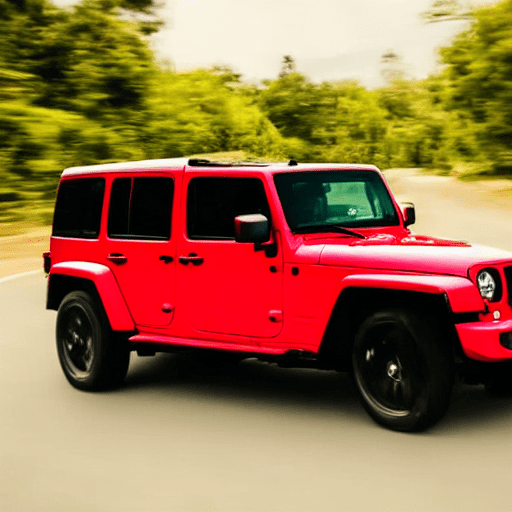}
    \includegraphics[width=0.115\textwidth]{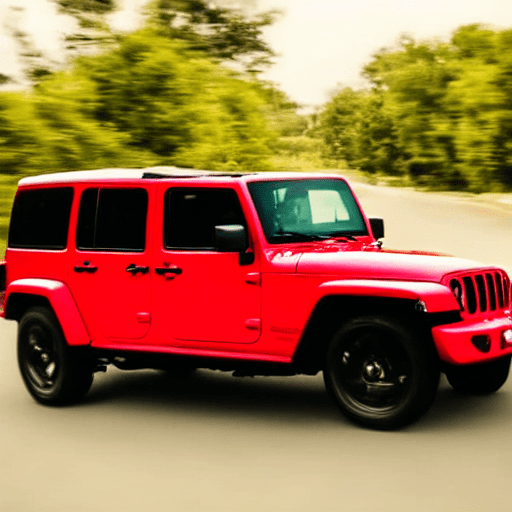}
    
    \makebox[0.115\textwidth]{A \colorbox{highlightcolor}{jeep car} is moving \colorbox{highlightcolor}{in the desert}.}\\
    \includegraphics[width=0.115\textwidth]{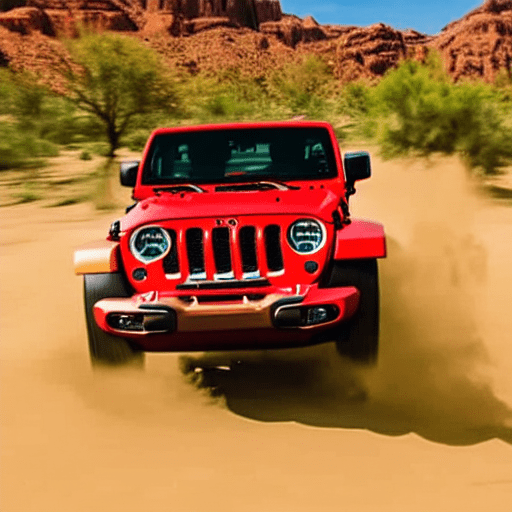}
    \includegraphics[width=0.115\textwidth]{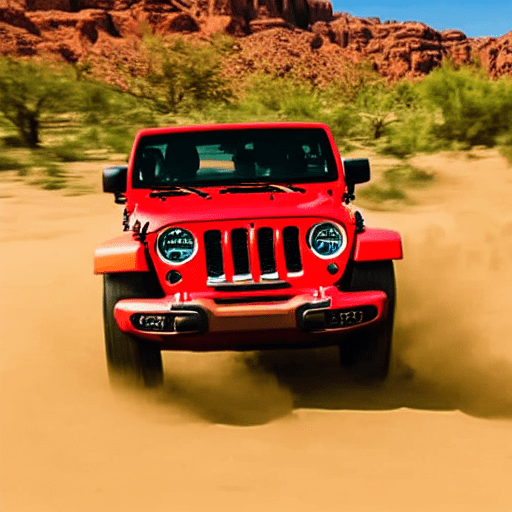}
    \includegraphics[width=0.115\textwidth]{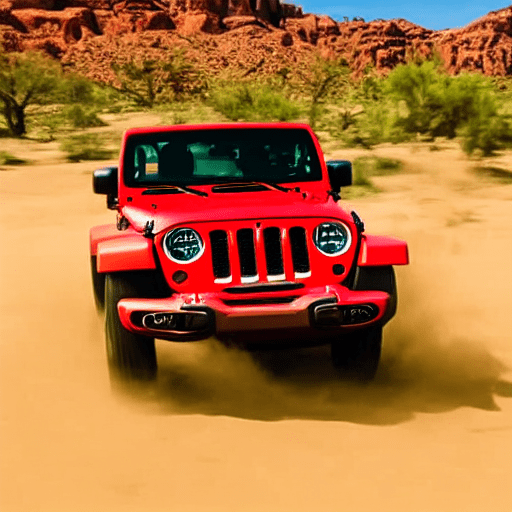}
    \includegraphics[width=0.115\textwidth]{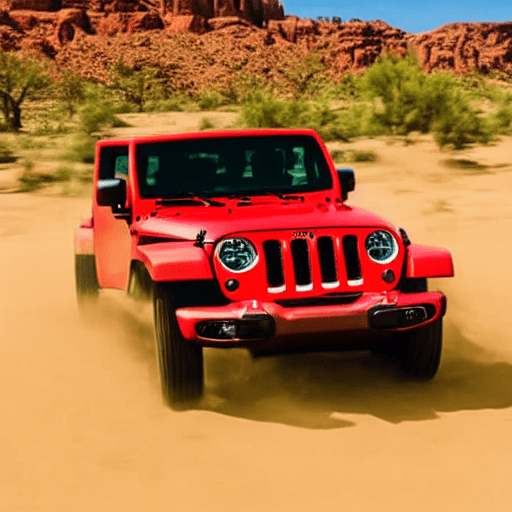}
    \includegraphics[width=0.115\textwidth]{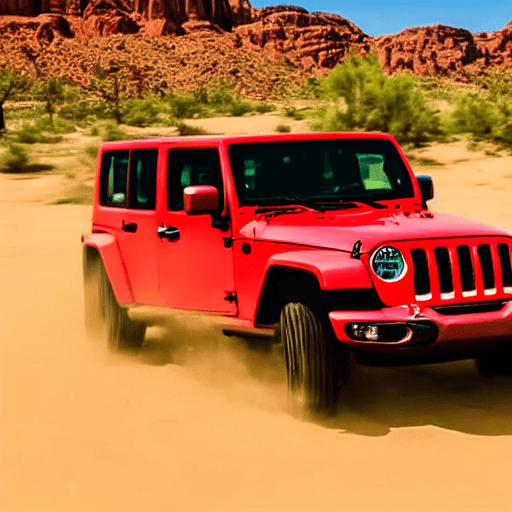}
    \includegraphics[width=0.115\textwidth]{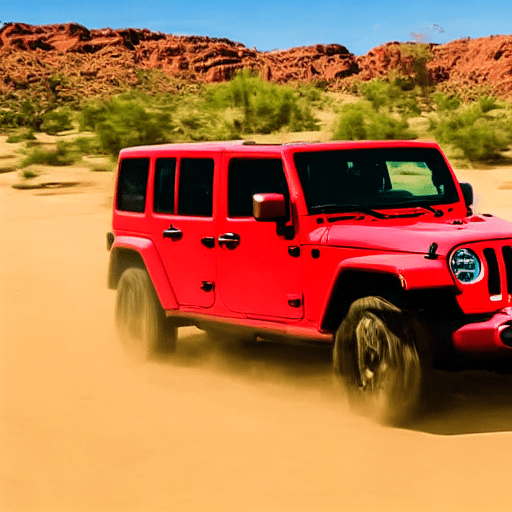}
    \includegraphics[width=0.115\textwidth]{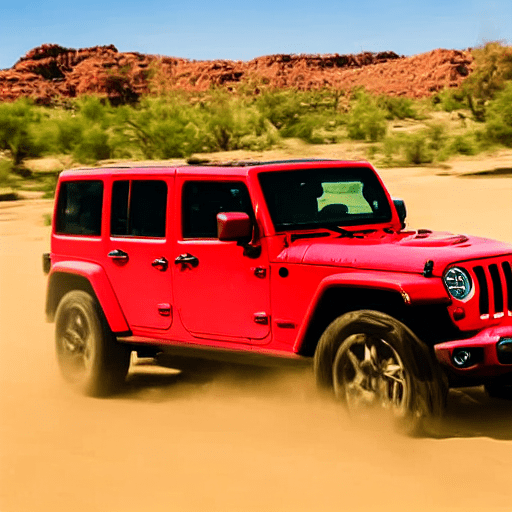}
    \includegraphics[width=0.115\textwidth]{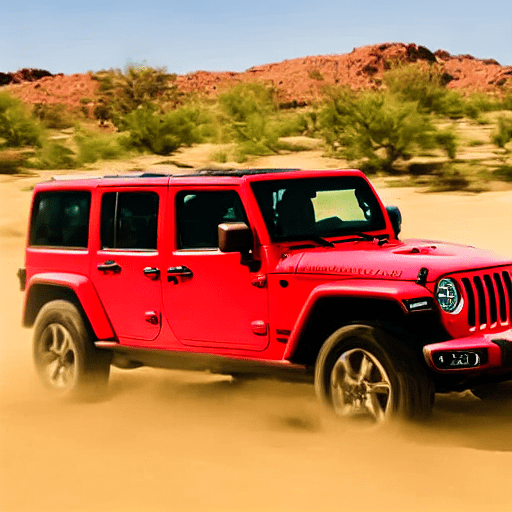}

    \caption{Videos editing results from various input videos and prompts. Our vid2vid-zero generates temporal consistent videos that align with the semantics of text prompts and faithfully to the original video. 
    }
    \label{fig:application_res}
    \end{figure*}
}

\newcommand{\figattnvis}{
    \begin{figure}[t]
    \centering

    \includegraphics[width=0.115\textwidth]{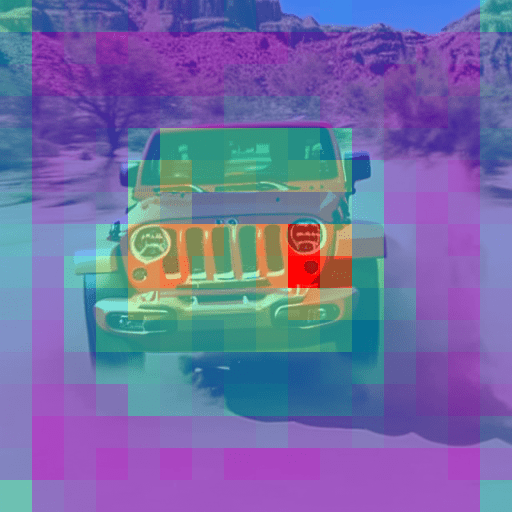}
    \includegraphics[width=0.115\textwidth]{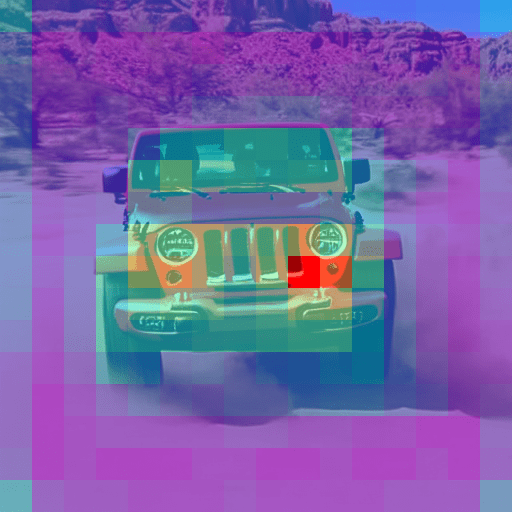}
    \includegraphics[width=0.115\textwidth]{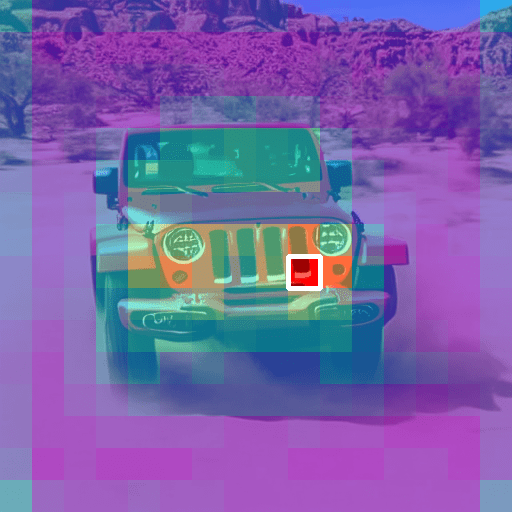}
    \includegraphics[width=0.115\textwidth]{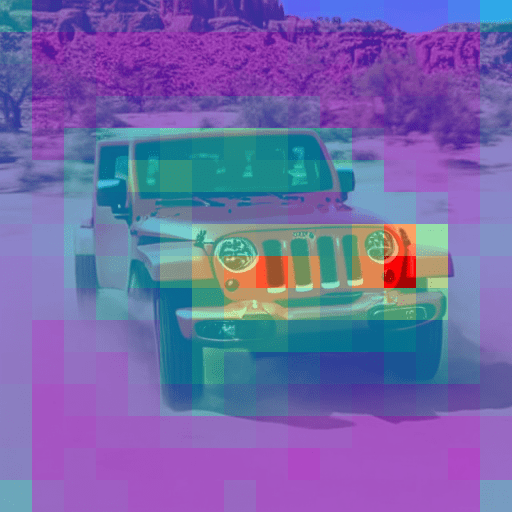}
    \includegraphics[width=0.115\textwidth]{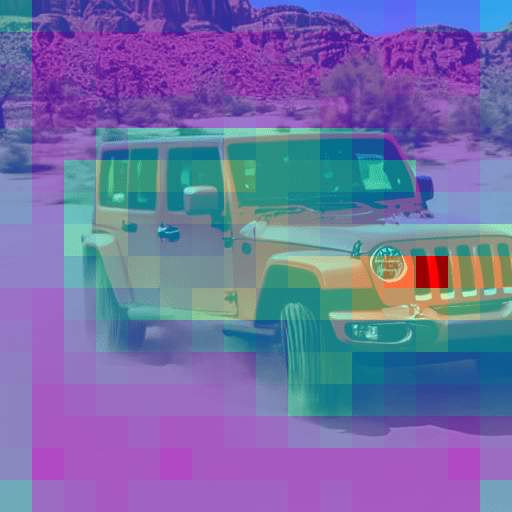}
    \includegraphics[width=0.115\textwidth]{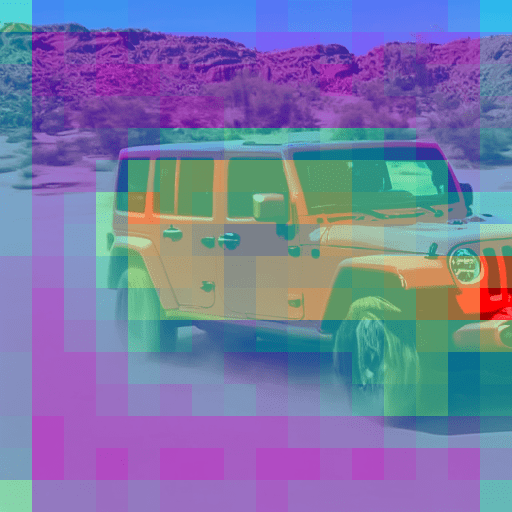}
    \includegraphics[width=0.115\textwidth]{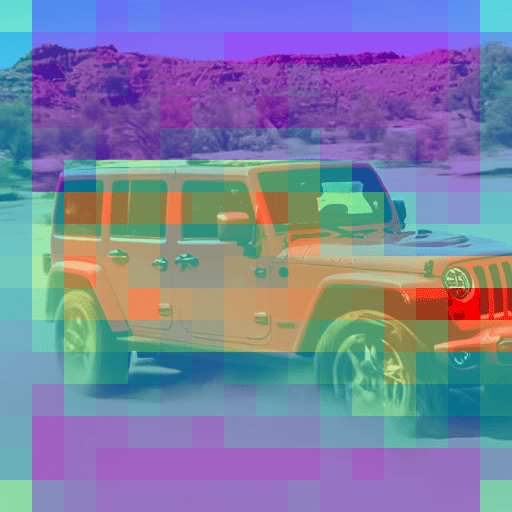}
    \includegraphics[width=0.115\textwidth]{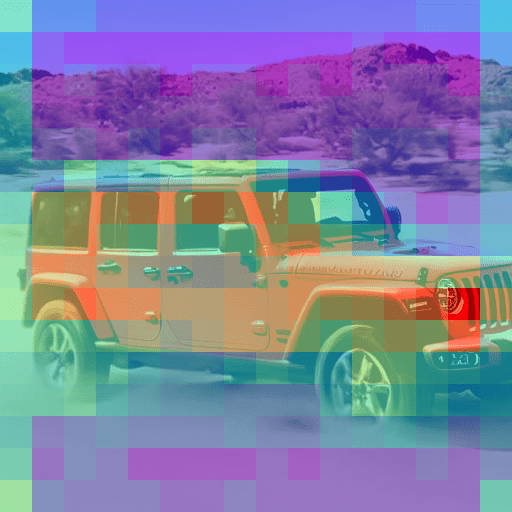} 
    \\
    
    \caption{Visualization of the spatial-temporal attention. See Fig.~\ref{fig:application_res} for the input and edited videos. The query is located near the right car light area in the third frame, highlighted by a white box.}
    \label{fig:attn_vis}
    \end{figure}
}

\newcommand{\figcomparsion}{
    \begin{figure}[t]
    \centering
    
    \makebox[0.11\textwidth]{\quad\quad 
    A horse is running on the beach.}
    \\
    
    \rotatebox{90}{\parbox{0.1\textwidth}{\centering 
    Input
    }}
    \includegraphics[width=0.11\textwidth]{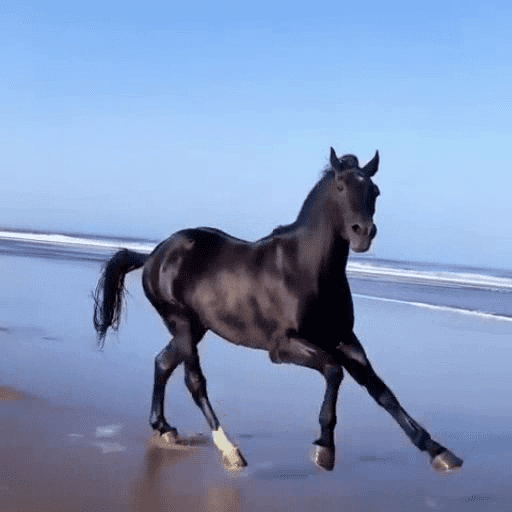}
    \includegraphics[width=0.11\textwidth]{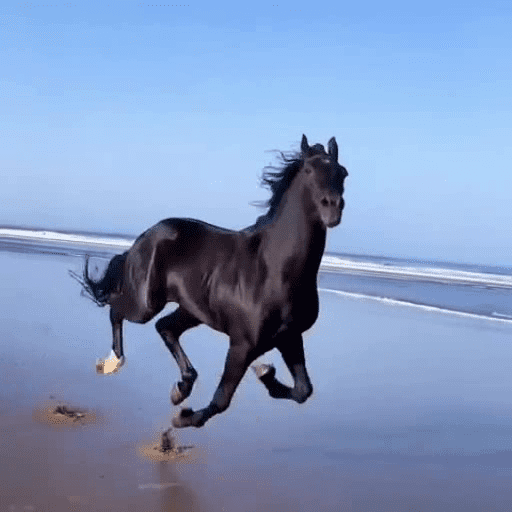}
    \includegraphics[width=0.11\textwidth]{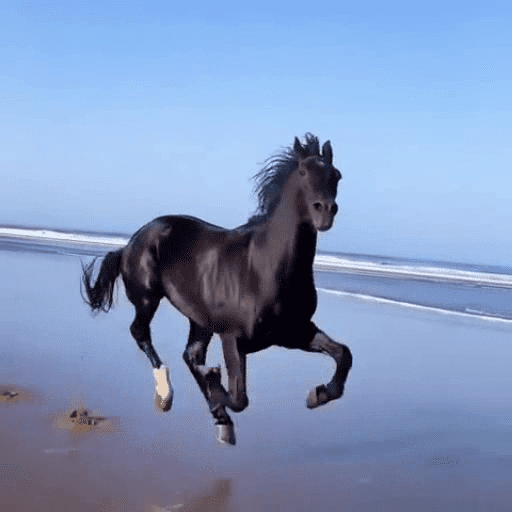}
    \includegraphics[width=0.11\textwidth]{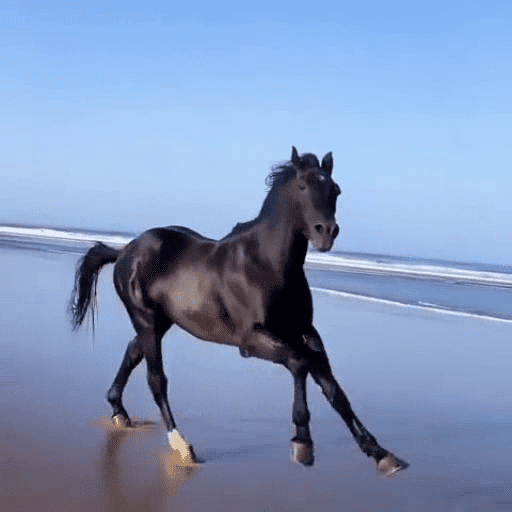}
    \\

    \makebox[0.11\textwidth]{\quad\quad A \colorbox{highlightcolor}{dog} is running on the beach.}
    \\

    \rotatebox{90}{\parbox{0.1\textwidth}{\centering 
    PnP
    }}
    \includegraphics[width=0.11\textwidth]{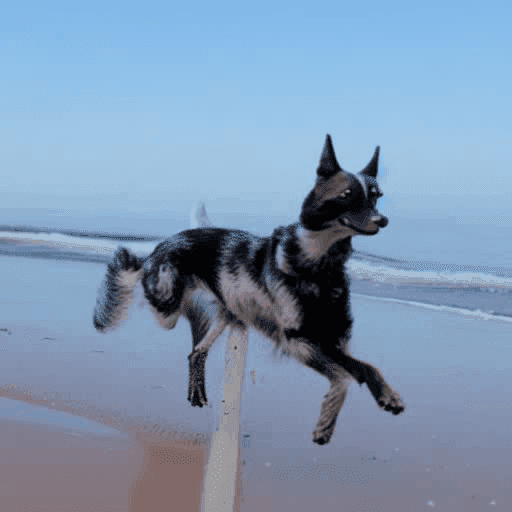}
    \includegraphics[width=0.11\textwidth]{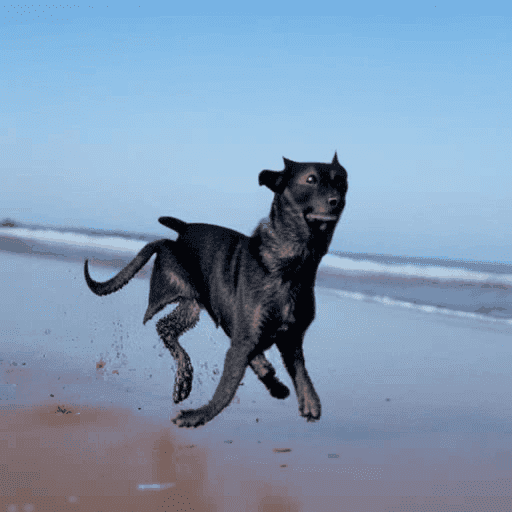}
    \includegraphics[width=0.11\textwidth]{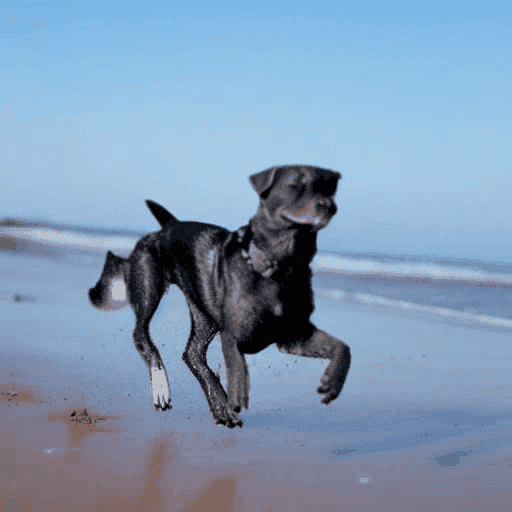}
    \includegraphics[width=0.11\textwidth]{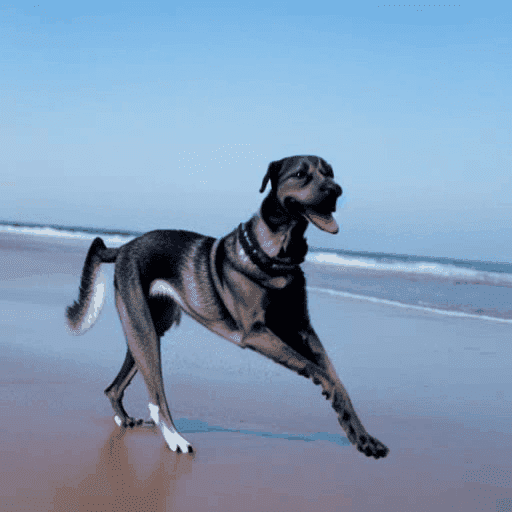}
    \\

    \rotatebox{90}{\parbox{0.1\textwidth}{\centering 
    TAV
    }}
    \includegraphics[width=0.11\textwidth]{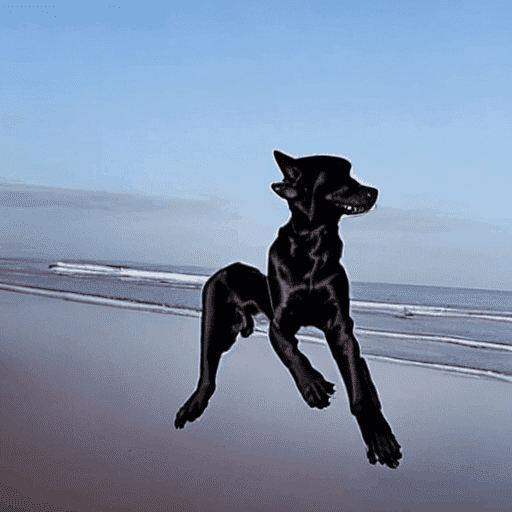}
    \includegraphics[width=0.11\textwidth]{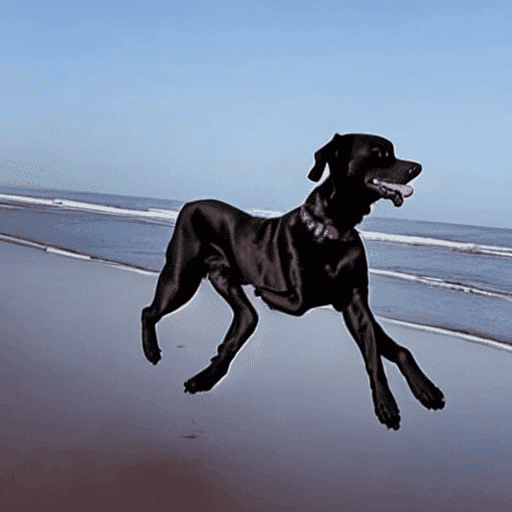}
    \includegraphics[width=0.11\textwidth]{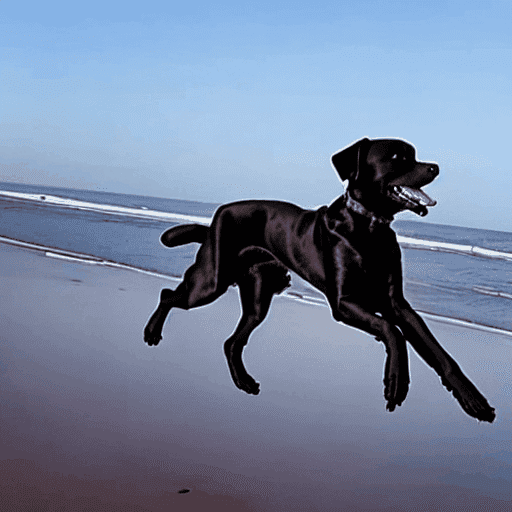}
    \includegraphics[width=0.11\textwidth]{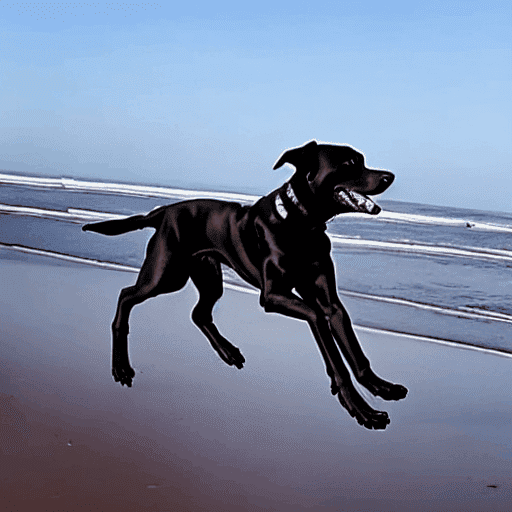}
    \\
    
    \rotatebox{90}{\parbox{0.1\textwidth}{\centering 
    Ours
    }}
    \includegraphics[width=0.11\textwidth]{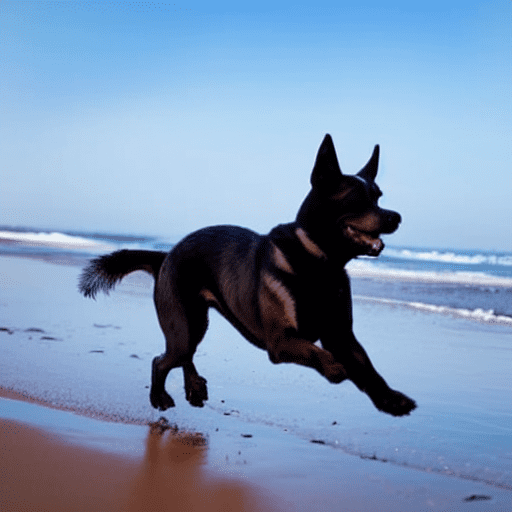}
    \includegraphics[width=0.11\textwidth]{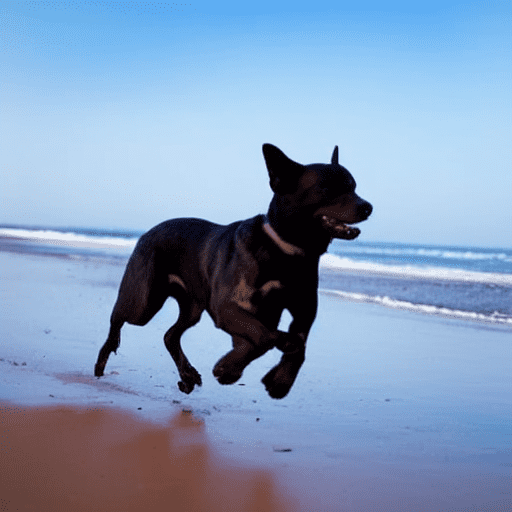}
    \includegraphics[width=0.11\textwidth]{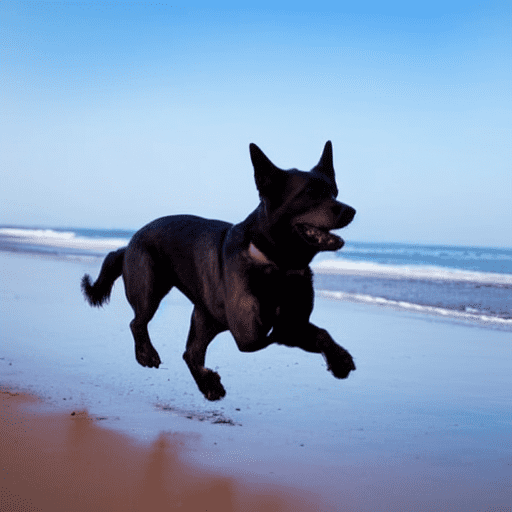}
    \includegraphics[width=0.11\textwidth]{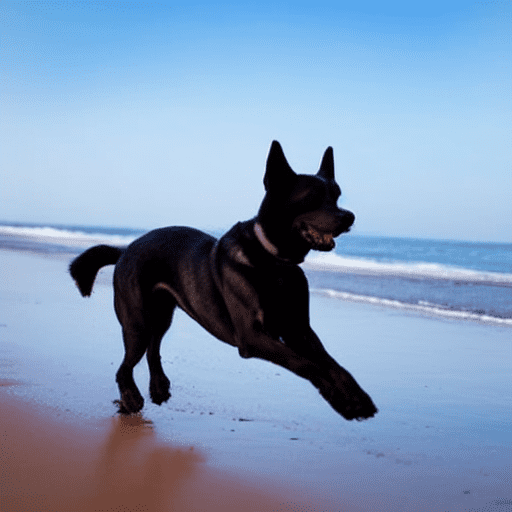}
    \\

    \caption{Comparison to other methods. Our vid2vid-zero achieves both temporal consistency and fidelity to the input video.}
    \label{fig:compare}
    \end{figure}
}

\newcommand{\figablation}{
    \begin{figure}[t]
    \centering
    
    \makebox[0.115\textwidth]{\quad\quad 
    A jeep car is moving on the road.}
    \\
    
    \rotatebox{90}{\parbox{0.1\textwidth}{\centering 
    Input \\ Video
    }}
    \includegraphics[width=0.1\textwidth]{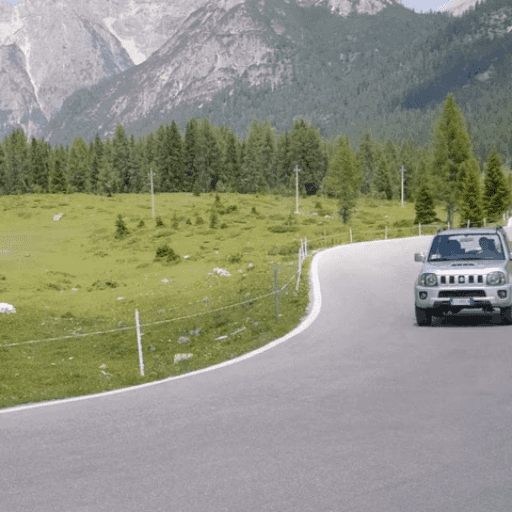}
    \includegraphics[width=0.1\textwidth]{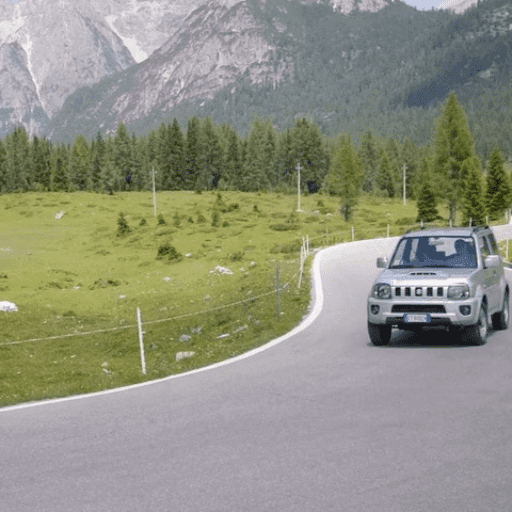}
    \includegraphics[width=0.1\textwidth]{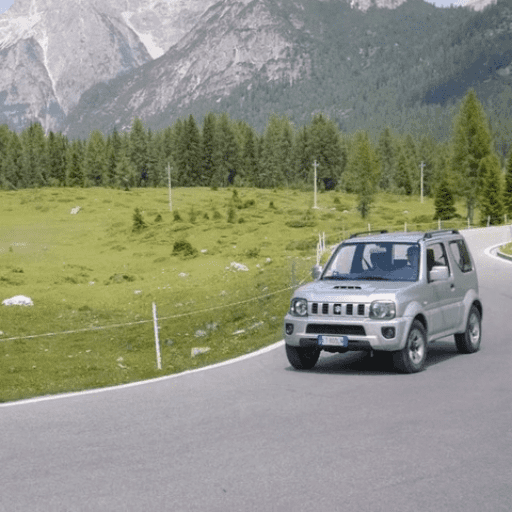}
    \includegraphics[width=0.1\textwidth]{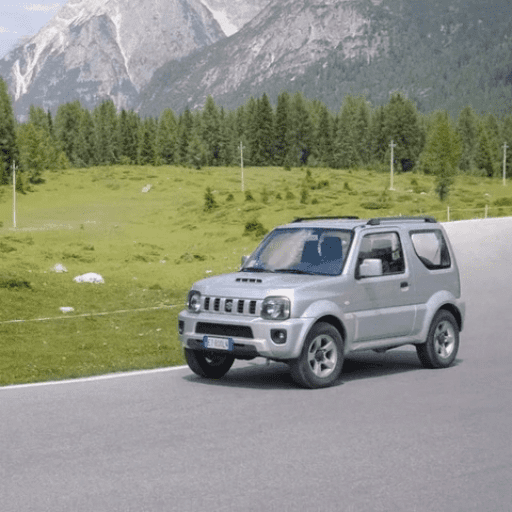}
    \\

    \makebox[0.115\textwidth]{\quad\quad A \colorbox{highlightcolor}{Porsche car} is moving \colorbox{highlightcolor}{in the desert}.}
    \\

    \rotatebox{90}{\parbox{0.1\textwidth}{\centering 
    w/o temporal
    }}
    \includegraphics[width=0.1\textwidth]{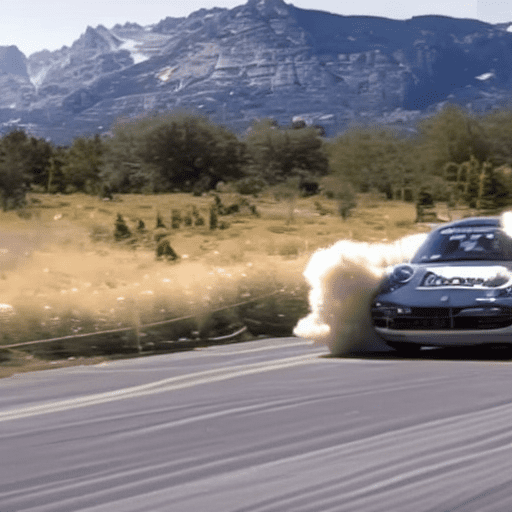}
    \includegraphics[width=0.1\textwidth]{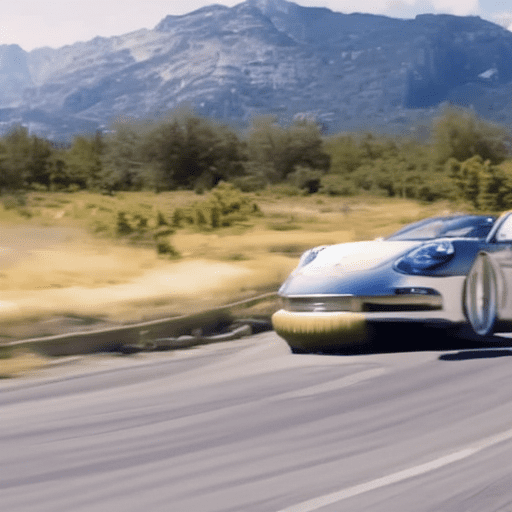}
    \includegraphics[width=0.1\textwidth]{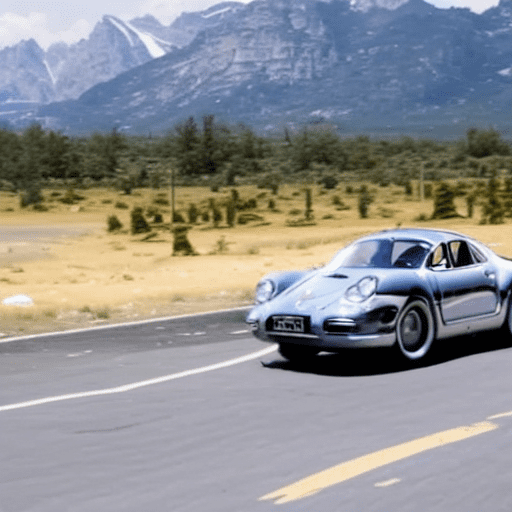}
    \includegraphics[width=0.1\textwidth]{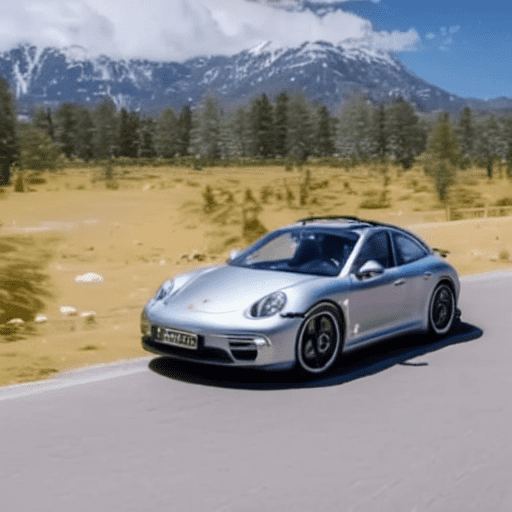}
    \\

    \rotatebox{90}{\parbox{0.1\textwidth}{\centering 
    w/o \\ spatial reg
    }}
    \includegraphics[width=0.1\textwidth]{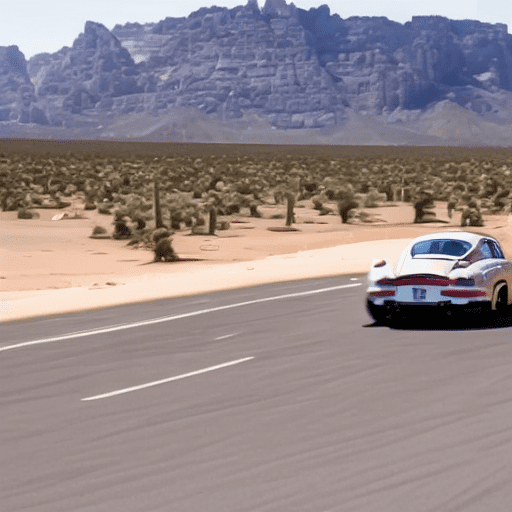}
    \includegraphics[width=0.1\textwidth]{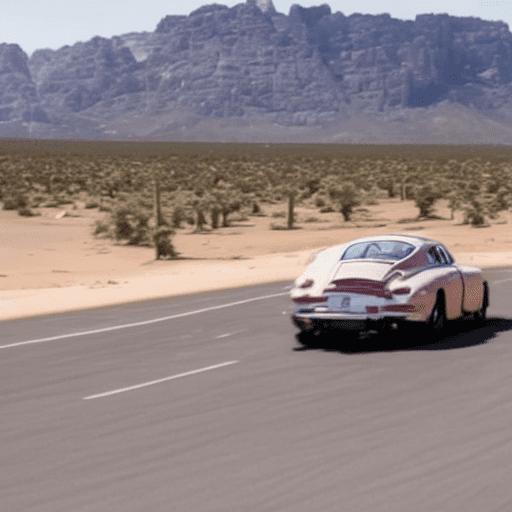}
    \includegraphics[width=0.1\textwidth]{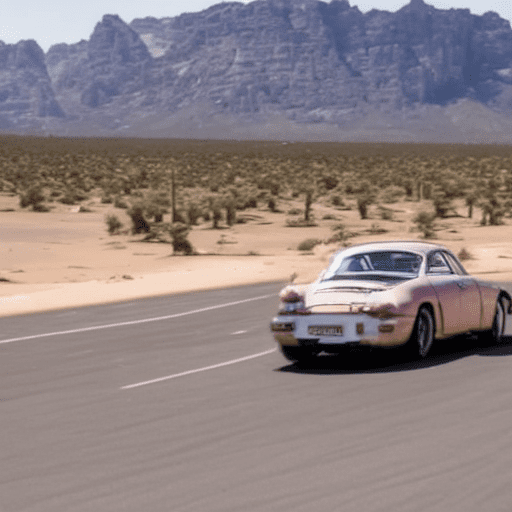}
    \includegraphics[width=0.1\textwidth]{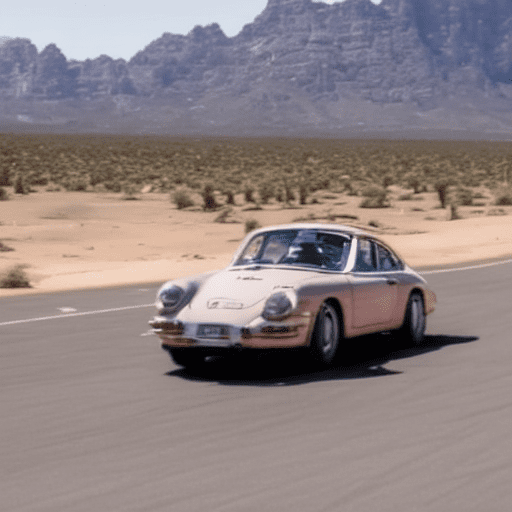}
    \\

    \rotatebox{90}{\parbox{0.1\textwidth}{\centering 
    w/o \\ null-inv
    }}
    \includegraphics[width=0.1\textwidth]{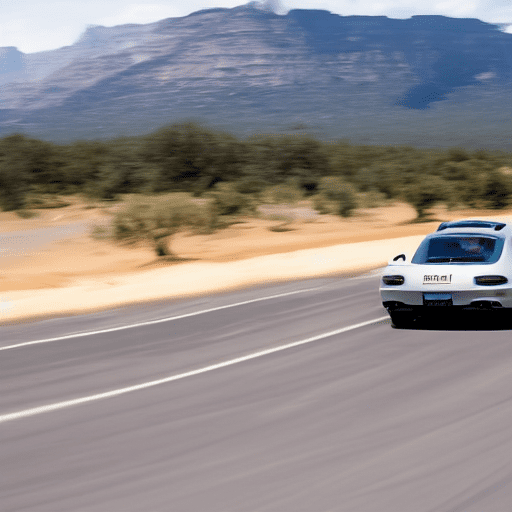}
    \includegraphics[width=0.1\textwidth]{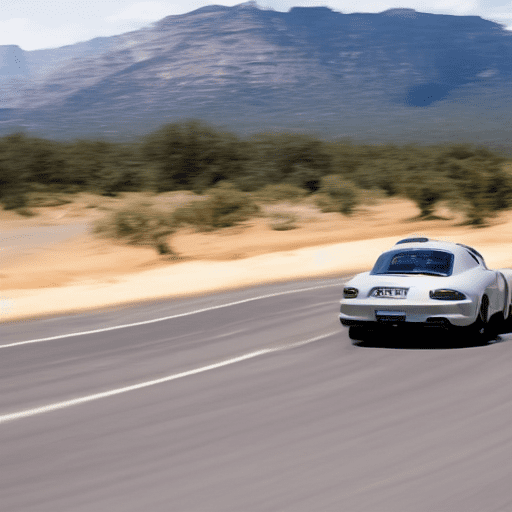}
    \includegraphics[width=0.1\textwidth]{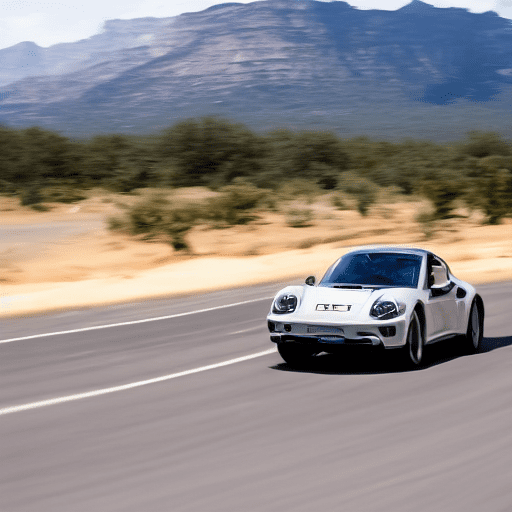}
    \includegraphics[width=0.1\textwidth]{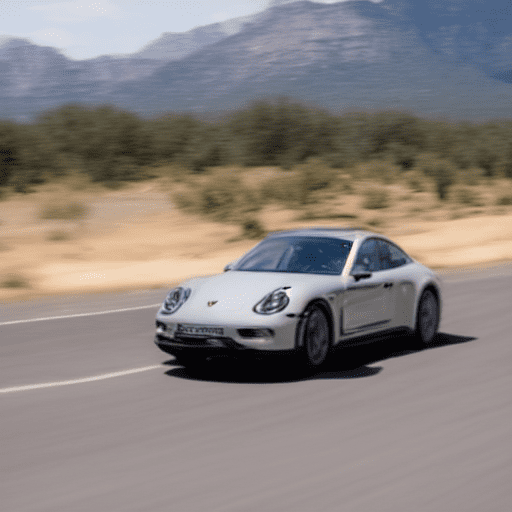}
    \\

    \rotatebox{90}{\parbox{0.1\textwidth}{\centering 
    ours \\
    full model \\
    }}
    \includegraphics[width=0.1\textwidth]{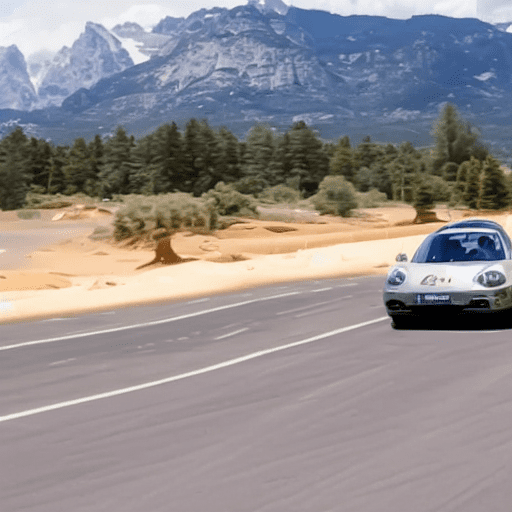}
    \includegraphics[width=0.1\textwidth]{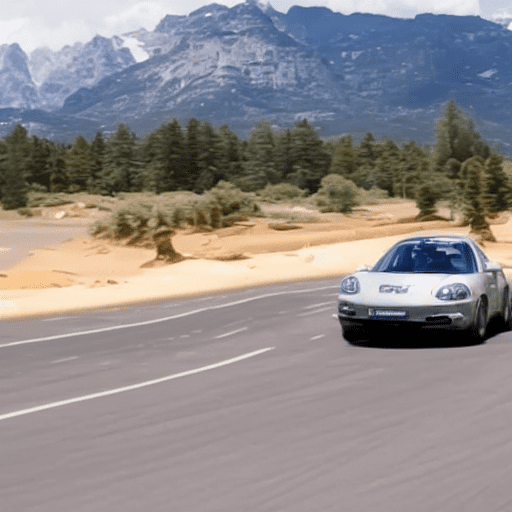}
    \includegraphics[width=0.1\textwidth]{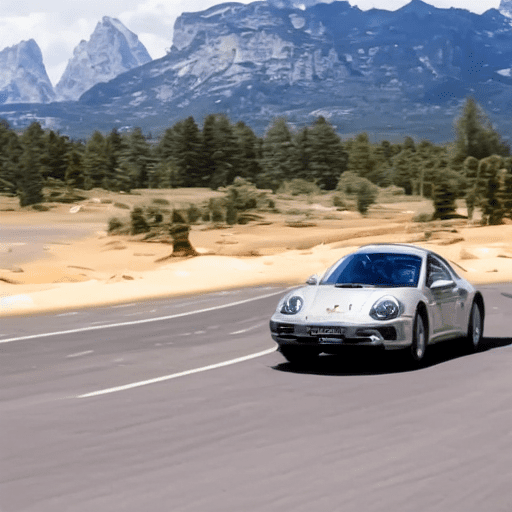}
    \includegraphics[width=0.1\textwidth]{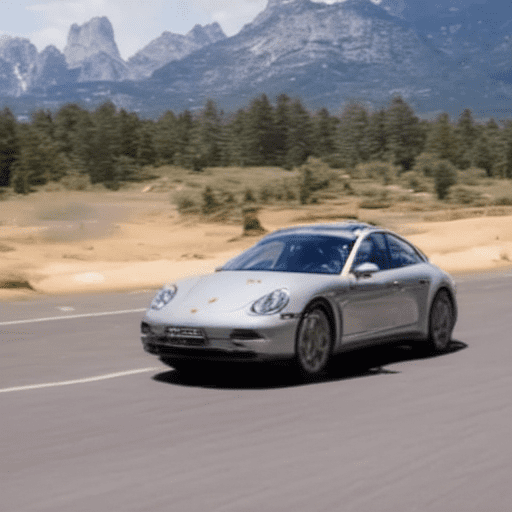}
    \\

    \makebox[0.115\textwidth]{\quad\quad [Same prompt. Ablations on temporal modeling]}
    \\
    
    \rotatebox{90}{\parbox{0.1\textwidth}{\centering 
    w/o \\ st-attn
    }}
    \includegraphics[width=0.1\textwidth]{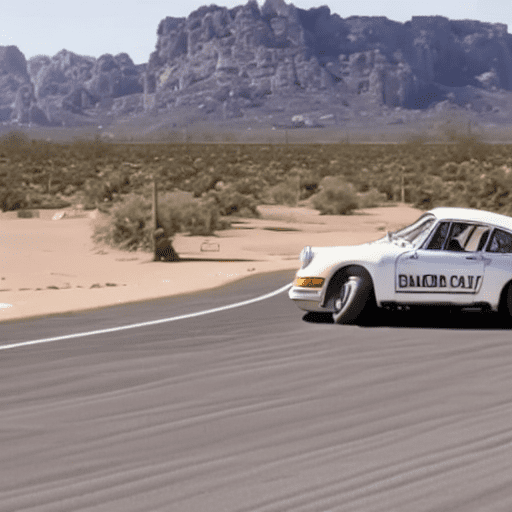}
    \includegraphics[width=0.1\textwidth]{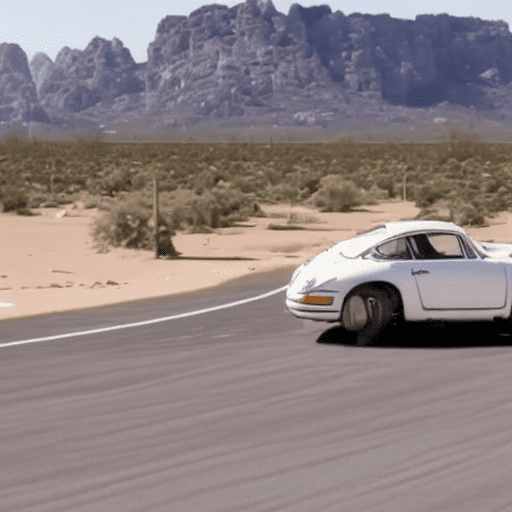}
    \includegraphics[width=0.1\textwidth]{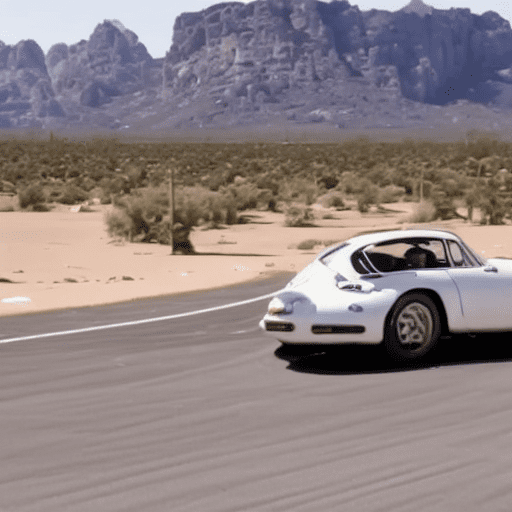}
    \includegraphics[width=0.1\textwidth]{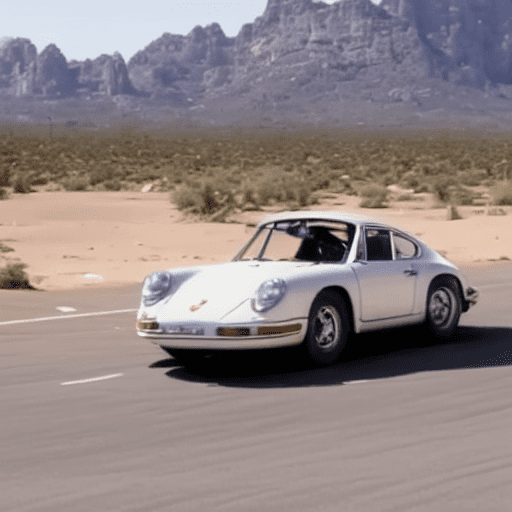}
    \\

    \rotatebox{90}{\parbox{0.1\textwidth}{\centering 
    temp-only attn\\ 
    }}
    \includegraphics[width=0.1\textwidth]{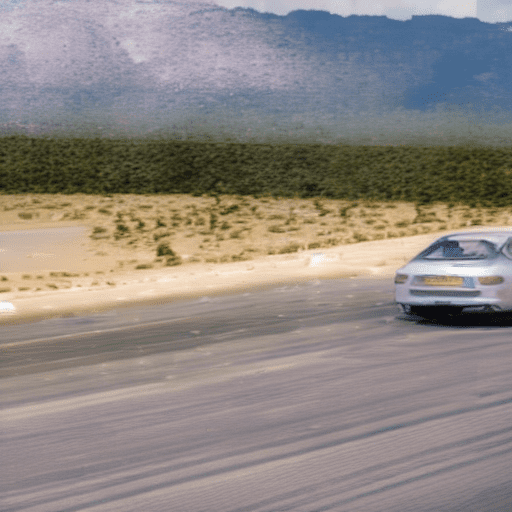}
    \includegraphics[width=0.1\textwidth]{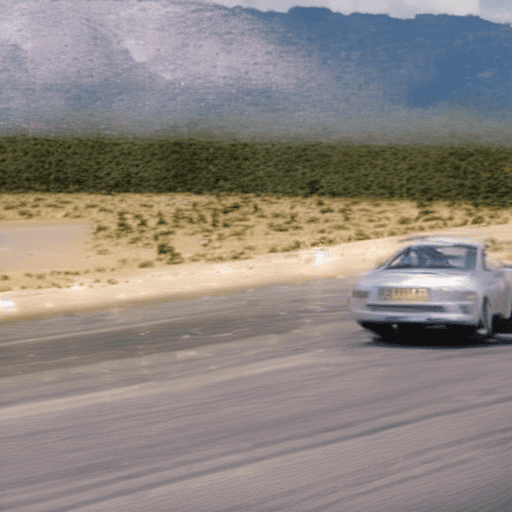}
    \includegraphics[width=0.1\textwidth]{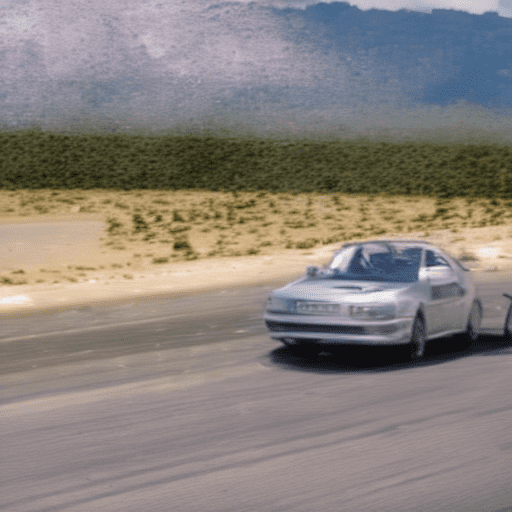}
    \includegraphics[width=0.1\textwidth]{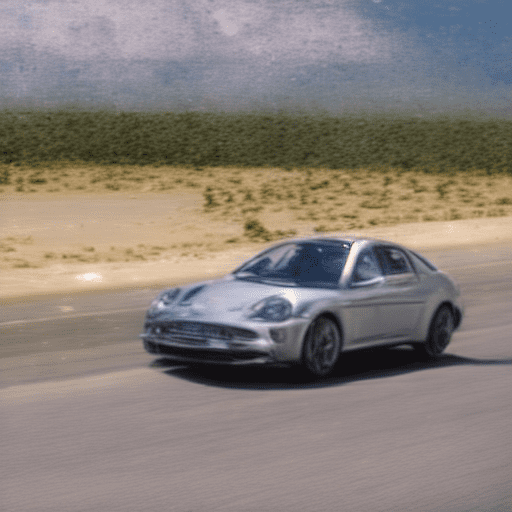}
    \\
    
    \caption{Ablations on the effectiveness of each component in vid2vid-zero (2nd $\sim$ 5th row), and the temporal modeling design (6th $\sim$ 7th row).}
    \label{fig:ablation}
    \end{figure}
}

\usepackage{blindtext}
\title{here title}
\let\oldtwocolumn\twocolumn
\renewcommand\twocolumn[1][]{%
    \oldtwocolumn[{#1}{
    \figteaser
    }]
}

\begin{document}
\title{Zero-Shot Video Editing Using Off-The-Shelf Image Diffusion Models}

\author{{
Wen~Wang, 
Yan Jiang, 
Kangyang~Xie, Zide~Liu, Hao~Chen, Yue~Cao, Xinlong~Wang, and~Chunhua~Shen}
\thanks{Wen Wang, Kangyang Xie, Zide Liu, Hao Chen, and Chunhua Shen are with
Zhejiang University, 
  China (e-mail: wwenxzy@zju.edu.cn
  haochen.cad@zju.edu.cn).
  } 
\thanks{Yan Jiang is with Harbin Engineering University, China (e-mail: y.jiang@hrbeu.edu.cn).}
\thanks{Xinlong Wang is with  Beijing Academy of Artificial Intelligence, Beijing 100085, China (e-mail: wangxinlong@baai.ac.cn). Yue Cao is an independent researcher (e-mail: caoyue10@gmail.com).} 
\thanks{
Hao Chen
is the corresponding author. First two authors contributed equally.
} 
\thanks{This work is supported by National Key R\&D Program of China (No.\  2022ZD0118700).}
}

\markboth{IEEE Trans. on Multimedia,~Vol.~XX, No.~XX, March~2023}%
{Shell \MakeLowercase{\textit{et al.}}: A Sample Article Using IEEEtran.cls for IEEE Journals}


\maketitle

\begin{abstract}

Large-scale text-to-image diffusion models achieve unprecedented success in image generation and editing. However, how to extend such success to video editing is unclear. Recent initial attempts at video editing require significant text-to-video data and computation resources for training, which is often not accessible. In this work, we propose vid2vid-zero, a simple yet effective method for zero-shot video editing. Our vid2vid-zero leverages off-the-shelf image diffusion models, and doesn't require training on any video. At the core of our method is a null-text inversion module for text-to-video alignment, a cross-frame modeling module for temporal consistency, and a spatial regularization module for fidelity to the original video. Without any training, we leverage the dynamic nature of the attention mechanism to enable bi-directional temporal modeling at test time. Experiments and analyses show promising results in editing attributes, subjects, places, etc., in real-world videos. Code is made available at \url{https://github.com/baaivision/vid2vid-zero}.

\end{abstract}



\section{Introduction}
\label{sec:intro}

\IEEEPARstart{T}{ext}-to-image diffusion models such as DALL·E 2~\cite{dalle2}, Imagen~\cite{imagen}, and Stable Diffusion~\cite{rombach2022high} are capable of generating unprecedented diverse and realistic images with complex objects and scenes, opening a new era of image generation. As an important application, image editing built on top of pre-trained diffusion models has also made significant progress~\cite{hertz2022prompt,couairon2022diffedit,kawar2022imagic,tumanyan2022plug,parmar2023zero}. 
These methods allow users to edit the input images through simple text prompts, and can achieve satisfying alignment with the target prompt and fidelity to the original image.

\IEEEpubidadjcol

However, it is still unclear how to extend such success to the video editing realm. Given the input video and text prompts, the text-driven video editing algorithm is required to output an edited video that satisfies (1) text-to-video alignment: the generated edited video should be aligned to the description of the text prompt; (2) fidelity to the original video: each frame of the edited video should be consistent in content with the corresponding frame of the original video; and (3) Quality: the generated video should be temporal consistent and in high quality.

Drawing on the success of the text-to-image diffusion model, one way is to build video editing algorithms on top of video diffusion models pre-trained on large-scale video datasets~\cite{molad2023dreamix,esser2023structure}. However, the training requires both significant amounts of paired text-video data and computational resources, which is often inaccessible. 
Instead, a recent work Tune-A-Video~\cite{wu2022tune} fine-tunes the pre-trained text-to-image diffusion model on a single video for video synthesis. While efficient in training, it still requires more than 70M parameters per video for fine-tuning. 
In addition, few-shot training updates the pre-trained weights and could lead to degradation in generation quality.

Different from these methods, we aim at performing zero-shot video editing using off-the-shelf image diffusion models, without training on any video. While it may appear easy to achieve video editing via frame-wise image editing techniques, severe flickering occurs even with the content-preserving DDIM inversion~\cite{song2020denoising} and cross-attention guidance~\cite{hertz2022prompt}, as shown in the first row in Fig.~\ref{fig:teaser}. 
The frame-wise image editing produces temporal inconsistent results when alternating the video style, due to the lack of temporal modeling.

To tackle this problem, we propose a simple yet effective pipeline, termed vid2vid-zero, for zero-shot video editing. Our method directly uses the pre-trained image diffusion models and is completely training-free. 
Specifically, it contains three major components, including a null-text inversion module for text-to-video alignment, a spatial regularization module for video-to-video fidelity, and a cross-frame modeling module for temporal consistency. 
To achieve the balance between effective training-free temporal modeling and reducing discrepancy to the sampling process of the pre-trained text-to-image diffusion model, a spatial-temporal attention module is proposed to model bi-directional temporal information at test time for video editing. Without bells and whistles, our method shows promising video editing results on real-world videos, as shown in the third row in Fig.~\ref{fig:teaser}.
Moreover, our vid2vid-zero can be easily combined with the customized image generation methods~\cite{ruiz2022dreambooth} to realize customized video editing, thanks to the zero-shot nature of our method.
To summarize, the contribution of this paper is listed as follows:
\begin{itemize}
    \item To our knowledge, we present the first zero-shot video editing method that directly uses the pre-trained text-to-image diffusion model, without any further training.
    \item 
    We leverage the dynamic nature of the attention mechanism and propose a simple yet effective dense spatial-temporal attention module that achieves bi-directional temporal modeling for video editing.
    \item 
    Experiments and analyses show promising results in editing attributes, subjects, places, \etc., in real-world videos.
\end{itemize}



\section{Related Work}
\subsection{Diffusion Models for Generation}
With the powerful capability of approximating data distribution, diffusion models have achieved unprecedented success in the image generation domain~\cite{ddpm, song2020denoising, song2021scorebased, song2020improved}. The magical generation fidelity and diversity outperform previous state-of-the-art generative models like Generative Adversarial Networks (GANs) ~\cite{Karras2019stylegan2, dhariwal2021diffusion, saharia2022palette, kingma2021variational}. Driven by pre-trained large language models~\cite{clip, raffel2020exploring}, text-to-image diffusion models can generate high fidelity images that are consistent with the text description~\cite{dalle2, imagen, rombach2022high, ramesh2021zero}. 

Compared to image generation, video generation is a more challenging problem due to its higher-dimensional complexity and the lack of high-quality datasets~\cite{villegas2022phenaki, hong2022cogvideo, wu2021godiva, nuwa}.
Encouraged by diffusion models' superiority in modeling complex higher-dimensional data and their success in image generation, recent literature attempts to extend diffusion models to the video generation task~\cite{ho2022video, imagen, esser2023structure, wu2022tune}.
As one of the pioneers, Video Diffusion Models~\cite{ho2022video} design a new architecture using 3D U-Net~\cite{3d-unet} with factorized space-time attention~\cite{arnab2021vivit, ho2019axial, bertasius2021space} for generating temporally-coherent results. Imagen Video~\cite{imagen} further scales up the Video Diffusion Model to achieve photo-realistic video generation. However, the difficulty of collecting large-scale text-video datasets and the expensive computational overheads are not feasible for most  researchers.

To this end, MagicVideo~\cite{zhou2022magicvideo} optimizes the efficiency by performing the diffusion process in the latent space. Leveraging a pre-trained text-to-image diffusion model, Make-A-Video~\cite{singer2022make}  performs video generation through a spatial-temporal decoder trained only on unlabeled video data. Recently, Tune-A-Video~\cite{wu2022tune} proposes a one-shot fine-tuning strategy on text-to-image pre-trained diffusion models for text-to-video generation. 

\subsection{Diffusion Models for Editing}

Besides the great success made in image generation, text-to-image diffusion models have also broken the dominance of the previous state-of-the-art models~\cite{ling2021editgan,jing2018stroke,jing2020dynamic,jing2022learning} in text-driven image editing~\cite{alaluf2021restyle,tov2021designing,richardson2021encoding}.
Building on top of a pre-trained text-to-image diffusion model, recent works~\cite{hertz2022prompt,couairon2022diffedit,kawar2022imagic,tumanyan2022plug,parmar2023zero} achieve impressive performance in text-driven image editing. 
Based on the key observation that the spatial layout and geometry information of generated images are retained in the text-image cross-attention map,
Prompt-to-Prompt~\cite{tumanyan2022plug} achieves a fine-grained control of the  spatial layout in the edited image by directly manipulating the cross-attention maps in the generating process. However, directly replacing the cross-attention maps may limit the editing scope of various motion changes. 
To tackle this problem, pix2pix-zero\cite{parmar2023zero} applies cross-attention guidance in generating process to ensure the structural information stays close to the input image but still has the flexibility to be changed under the guidance of the input prompt. Furthermore, they propose to compute the edit direction based on multiple sentences in the text embedding space which allows for more robust and better performance in text-driven image editing.
With similar motivation, Plug-and-Play\cite{tumanyan2022plug} achieves the state-of-the-art text-driven image editing performance with satisfying semantic and spatial correspondence by replacing spatial features in generating process with extracted corresponding counterparts of the input image.

While text-to-image diffusion models have made great success in image editing, few explorations have been made for extending them to the video editing realm.
Dreamix~\cite{molad2023dreamix} pioneers the application of a diffusion-based model for text-driven video editing. They fine-tune the video diffusion model~\cite{ho2022video} on both original videos and unordered frames set with a mixed objective to improve the performance of motion edits and temporal modeling. GEN-1~\cite{esser2023structure} achieves impressive video-editing performance under the guidance of the input prompt without fine-tuning on individual input videos. However, it requires training a latent video diffusion model conditioned on depth maps and input prompts to maintain the structure consistency. 
Different from these methods that are built on top of large-scale trained video diffusion models, we completely bypass the requirements of expensive computational resources and the significant amount of text-video training data, and instead achieve zero-shot video editing using the off-the-shelf image diffusion models.

Several concurrent works~\cite{liu2023video,shin2023edit,qi2023fatezero,ceylan2023pix2video} also make attempts to tackle the video editing problem. However, they either require fine-tuning the image diffusion model on the input video~\cite{liu2023video,shin2023edit} or rely on a video diffusion model to perform zero-shot video editing~\cite{qi2023fatezero}.
Probably the most similar work to our vid2vid-zero is Pix2Video~\cite{ceylan2023pix2video}, which relies on a pre-trained structure-guided (\eg, depth) diffusion model and utilizes features from the previous and the first edited frames to edit the current frame. Differently, we do not assume the image diffusion model is structure-guided and can take advantage of the bi-directional temporal modeling for temporal consistency.


\section{Method}

In text-driven video editing, the user provides an input video $X_0=\left\{x_0^i \mid i \in[1, F]\right\}$ with $F$ frames, with the corresponding text description (source prompt) $c$, and then queries the video editing algorithm with a target prompt $\hat{c}$ to produce the edited video. 
To tackle this problem, we propose a simple yet effective method vid2vid-zero, that edits video in a zero-shot manner using the publicly available text-to-image diffusion model, as shown in Figure~\ref{fig:method}. 
Specifically, our method contains three major components, a video inversion module for text-to-video alignment (Sec.~\ref{subsec:inversion}), a spatial regularization module for video-to-video fidelity (Sec.~\ref{subsec:spatial}), and a cross-frame modeling module for temporal consistency (Sec.~\ref{subsec:temporal}).

\figmethod

\begin{algorithm}[t!]
\caption{vid2vid-zero algorithm}
\label{alg:vid2vid-zero}
\begin{algorithmic}

\small 

\State \textbf{Input:} $X_0$:  \text{input video} \\
$c$: \text{source prompt} \\
$\hat{c}$: \text{target prompt} \\
$\tau_{null}, \tau_{M}$: \text{
injection thresholds} \\
\\
 $\triangleright$ Real Video Inversion \\
 $\left\{X_t^{\text{inv}}\right\}_{t=1}^T = \Call{DDIM-inv}{X_{0}}$ \\
 $\left\{\varnothing_t\right\}_{t=1}^T = \Call{Null-Opt}{\left\{X_t^{\text{inv}}\right\}_{t=1}^T, c}$ \\
 \\
 $\triangleright$ Computing reference cross-attention maps
\State ${X}_T \gets X^{\text{inv}}_{T}$ 
\For{$t = T...1$}

    \State $\triangleright$ null-text embedding is injected
    \State $z_{t}, M_{t} \gets \hat{\epsilon}_{\theta}\left(X_t, t, c; \varnothing_t \right)$
    \State $X_{t-1}=\Call{DDIM-samp}{X_t, z_{t}, t}$

\EndFor

\\
\State $\triangleright$ Editing with guidance
\State $\hat{X}_T \gets X^{\text{inv}}_{T}$ 
\For{$t = T...1$}
    \State \text{if} $t < \tau_{M}$ \text{then} $M_t \gets \emptyset$  
       $\triangleright$ without injection
    \State \text{if} $t < \tau_{null}$ \text{then} $\varnothing_{t} \gets \emptyset$
    \State $\hat{z}_{t}, \_\_ \gets \hat{\epsilon}_\theta\left(\hat{X}_t, t, \hat{c} ; \varnothing_t, M_t \right)$ 
    \State $\hat{X}_{t-1}=\Call{DDIM-samp}{\hat{X}_t, \hat{z}_{t}, t}$

\EndFor
\State {\bf Output:} $\hat{X}_0$: edited video 
\end{algorithmic}
\end{algorithm}

\subsection{Real Video Inversion}
\label{subsec:inversion}

\paragraph{DDIM Inversion.}
A common practice in image editing is to find the variables in the latent space that corresponds to the image. Afterward, image editing can be achieved by finding editing directions in the latent space. Motivated by their success in image editing~\cite{mokady2022null,parmar2023zero,parmar2023zero}, we first inverse each frame in the input video to the noise space, through the commonly used deterministic inversion method, DDIM inversion~\cite{song2020denoising}. The inversion trajectory over $T$ timesteps can be denoted as $\left\{X_t^{\text {inv }}\right\}_{t=1}^T$.

\paragraph{Null-text Optimization.}
Although using DDIM inversion preserves the information of each frame in the video, the obtained latent noise may not be aligned to the user-provided text description $c$. In other words, when sampling with the latent $X_T^{\text {inv}}$ and the source prompt, the reconstructed video may be significantly different from the original video. 
To solve the above problem, we resort to prompt-tuning~\cite{liu2023pre} to learn a soft text embedding that aligns with the video content. Although more sophisticated methods in prompt-tuning can be used, we find that optimizing the null-text embedding~\cite{mokady2022null} can achieve promising text-to-video alignment. Taking the DDIM inversion trajectory $\left\{X_t^{\text {inv }}\right\}_{t=1}^T$ and the source prompt $c$ as input, the optimized null-text embedding trajectory can be written as $\left\{\varnothing_t\right\}_{t=1}^T$. 
Specifically, the null-text embedding $\varnothing_t$ is updated according to the following equation:
\begin{equation}
\min _{\varnothing_t}\left\|X_{t-1}^{\text{inv}}-{f}_\theta\left(\overline{X}_t, t, c; \varnothing_t \right)\right\|_2^2,
\end{equation}
where ${f}_\theta$ denotes applying DDIM sampling on top of the noise predicted by pre-trained image diffusion model $\epsilon_{\theta}$, $\overline{X}_t$ is the sampled latent code during null-text inversion. 
In practice, we share the same null-text embedding for different video frames, in order to preserve consistent information across video frames.

\figattntypes

\subsection{Temporal Modeling}
\label{subsec:temporal}
Temporal modeling is essential for video editing. The dynamic computation nature makes the attention mechanism a suitable choice for temporal modeling at test time. To this end, Sparse-Causal Attention~\cite{wu2022tune} (SC-Attn) is recently proposed to replace self-attention in pre-trained diffusion models, to capture the causal temporal information required for video generation. As shown in Fig.~\ref{fig:attn_type}(b), it queries tokens in both the previous and the first frames for temporal modeling. 
While effective for generation, it is not optimal for the video editing task, as bi-directional temporal modeling is necessary to ensure temporally-coherent video editing results. An alternative choice for bi-directional temporal modeling is the temporal-only attention~\cite{CogVideo,singer2022make}, as shown in Fig.~\ref{fig:attn_type}(c). However, it completely abandons spatial modeling, leading to a significant discrepancy with the self-attention in pre-trained image diffusion models (Fig.~\ref{fig:attn_type}(a)). Our experiments in Sec.~\ref{subsec:exps} indicate that this discrepancy leads to severe degradation in editing performance. 

To reach a trade-off between bi-directional temporal modeling and alignment to the sampling process of the pre-trained image diffusion model, we propose spatial-temporal attention (ST-Attn) for video editing. As shown in Fig.~\ref{fig:attn_type}(d), each frame $x_{i}$ attends to all frames in the video. Specifically, query features are computed from all spatial features in the query frame $x_{i}$ while key and value features are computed from all spatial features across all frames $x_{1:T}$. 

Mathematically, the query, key, and value in the proposed ST-Attn can be written as:
$Q=W^Q {x}_i$, $K=W^K {x}_{1:T}$, and $V=W^V {x}_{1:T}$, respectively.
Here, $W^Q$, $W^K$, and $W^V$ are the pre-trained projection weights in the self-attention layers, and are shared by all tokens in different spatial and temporal locations.

Attending both previous frames and future frames with the proposed spatial-temporal attention provides bi-directional temporal modeling capability, while querying with spatial features across locations alleviates the discrepancy to the sampling process of pre-trained image diffusion models.

As shown in Fig.~\ref{fig:method}(c), we replace the pre-trained self-attention modules with cross-frame attention that shares exactly the same weights. In practice, We find that using ST-Attn at the beginning of the down-sampling, middle, and up-sampling block, and replacing the other self-attention layers with SC-Attn already works well for bi-directional temporal modeling. Besides, the original 2D residual block is inflated to 3D by copying the weights to each frame, to enable the inference on video inputs. We denote the model with updated U-Net blocks as $\hat{\epsilon}_{\theta}$.

\subsection{Spatial Regularization}
\label{subsec:spatial}
Maintaining fidelity to the original input video is an important aspect of the video editing task. To this end, we introduce spatial regularization to the editing process. 
The video inversion in Sec.~\ref{subsec:inversion} can largely ensure the reconstruction of the original video when sampling with the inversion results and the source prompt. And the cross-attn maps generated during reconstruction contain the spatial information of the original video~\cite{hertz2022prompt}. For this reason, we can use the cross-attention maps as a spatial regularization and force the model to focus on the prompt-related areas via attention map injection.
Formally, the noise prediction process can be written as:
\begin{equation}
\hat{z}_t = \hat{\epsilon}_\theta\left(\hat{X}_t, t, \hat{c} ; \varnothing_t, M_t\right),
\end{equation}
where $M_t$ is the attention mask produced during the reconstruction of the input video. 
Here, we use ${\epsilon}_\theta\left(\cdot \, \boldsymbol{;} \, \varnothing_t, M_t\right)$ to denote the denoising steps with injected cross-attention map $M_t$ and null-text embedding $\varnothing_t$.
Thanks to our zero-shot temporal modeling methods in Sec.~\ref{subsec:temporal}, the guidance cross-attention mask is temporal-aware, thus preventing sudden changes in cross-attention maps over different frames.


\figresults

\section{Experiments}

\subsection{Implementation Details}
In experiments, we implement our method on top of the latent diffusion models~\cite{rombach2022high} and use the publicly available Stable Diffusion model weights\footnote{https://huggingface.co/CompVis/stable-diffusion-v1-4} by default. More results with different text-to-image diffusion model weights are presented in the Appendix.  Following Tune-A-Video~\cite{wu2022tune}, each video contains 8 frames in 512$\times$512 resolution by default. During inference, we use DDIM sampler~\cite{song2020denoising} with 50 steps and classifier-free guidance~\cite{ho2022classifier} with a guidance scale of $7.5$.
For null-text inversion, we follow the hyper-parameters in ~\cite{mokady2022null}, except that we set the inner step as 1, which we found already works well.
The cross-attention injection threshold and the null-text embedding injection threshold are set as 0.8 and 0.5, respectively.

\subsection{Main Results}
We showcase the effectiveness of our zero-shot video editing method in Fig.~\ref{fig:application_res}. Specifically, two sets of video examples are included, whose text descriptions are ``a man is running" and ``A car is moving on the road", respectively. More demonstrations are provided in the Appendix.

\paragraph{Editing Style.}
The style of a video is reflected by the global spatial and temporal characteristics of the video. In the 2nd row in Fig.~\ref{fig:application_res}, we add ``anime style" to the input video. As can be seen, our method is capable of transforming all frames in the video to the target style, without alternating the semantics of the original video. 

\paragraph{Editing Attributes.}
In the 3rd and 7th rows in Fig.~\ref{fig:application_res}, we demonstrate the ability of vid2vid-zero to edit attributes. For example, turning a young man into an old man and changing the model of the car. As can be seen, the edits are highly consistent with the target prompt. What's more, they also maintain the other attributes of the original video, like the color and the pose of the car in the 7th row.

\paragraph{Editing Background.}
Rows 3, 4, 6, and 7 of Fig.~\ref{fig:application_res} show the results of vid2vid-zero on editing background. As can be seen, it successfully turns the backgrounds to beach, complex urban street scenes, and the desert. These results demonstrate the ability of vid2vid-zero to retain the generative power of the pre-trained Stable-Diffusion, enabling creative video editing creations in a zero-shot manner. we found that when editing the background, the model will also change accordingly to make the overall effect more realistic. As row 6 in Fig.~\ref{fig:application_res}, when changing the prompt to ``A car is moving on the snow", the model also adds snow on the top of the car according to the scene.

\paragraph{Replacing Subjects.}    
To demonstrate the effectiveness of our method on editing subjects, we turn the man into Stephen Curry in the 4th row of Fig.~\ref{fig:application_res}, and replace the horse with the dog in the 4th row of Fig.~\ref{fig:compare}. As can be seen, our editing results not only align well with the text description but also maintain fidelity to the original videos. 
What's more, our method can edit multiple properties at the same time, for example, replacing subjects (replacing the man with Curry) and editing background (changing the background to Time Square) at the same time.

\subsection{Analysis}
\label{subsec:exps}
\figablation
\paragraph{Ablations on Each Module.}
We conduct ablation studies by isolating each component in vid2vid-zero, as shown in the 2nd to 4th row in Fig.~\ref{fig:ablation}. We have the following observation. Firstly, without the proposed temporal modeling, both the foreground car and the background mountain become inconsistent over time. Secondly, without spatial attention guidance, the edited video became less faithful to the input video. For example, the color of the car changes, and the background trees are missing. Thirdly, without null-text inversion, the background mountain and trees become blurry, since the optimized null-text embedding contains fine-grained details that align with the input video. Lastly, the above three components are complementary to each other, and combining them together gives rise to the best video editing result.

\figcomparsion
\paragraph{Ablations on Temporal Attention Modeling.}
Effective temporal modeling is the key to achieving zero-shot video editing. To this end, we perform further ablation experiments on the temporal modeling design. First, we consider removing the dense spatial-temporal attention that enables bidirectional temporal modeling and replacing it with Sparse-Causal Attention. As shown in the 6th row of Fig.~\ref{fig:ablation}, the car in the first few frames of the video is deformed. This is mainly because SC-Attention over-emphasizes the previous frames of the video, and the editing errors in the first few video frames propagate to the subsequent frames, leading to serious artifacts.
Besides, we also present the results of replacing Dense-attn with Temporal-only Attention, which only focuses on temporal modeling while ignoring other spatial locations. As can be seen in the last row of Fig.~\ref{fig:ablation}, Temporal-only Attention can also alleviate the error propagation caused by the first few frames. However, it has a large gap compared with the pre-training self-attn, which on contrary only focuses on spatial location.  As a result, there is a large distribution gap between the edited video frames and the output space of the pre-training text-to-image diffusion model, as indicated by the blurry video frames.

\paragraph{Spatial-Temporal Attention Visualization.}
\figattnvis
To gain a better understanding of the behavior of the proposed dense spatial-temporal attention, we visualize the attention maps in Fig.~\ref{fig:attn_vis}. The target prompt is ``A jeep car is moving in the desert", and the corresponding input video and edited video are shown in Fig.~\ref{fig:application_res}. The query is located at the right car light region in the third frame, highlighted by the white box. As can be seen, it successfully attends the car light areas in both previous and future frames, which demonstrates the effectiveness of our method in capturing bidirectional temporal and spatial information.

\subsection{Comparison}

\paragraph{Compared Methods.}
\textit{Tune-A-Video (TAV)}\cite{wu2022tune} realizes video generation via fine-tuning a pre-trained text-to-image diffusion model on a single video, and has shown potential applications in subject replacement, background change, attribute modification, and style transfer.
\textit{Plug-and-Play (PnP)}~\cite{tumanyan2022plug} is a state-of-the-art image editing method, which achieves high-quality image editing via feature replacement. We use PnP to edit the input video frame-by-frame and concatenate them to obtain the edited video.

\paragraph{Qualitative Comparison.}
The results of different methods on editing subjects are shown in Fig~\ref{fig:compare}. As can be seen, each frame produced by PnP shares similar poses to the corresponding input frame. However, it suffers from flickering effects like the sudden appearance of white bands in the first column and the deformation and repeated tails in the 2nd and 3rd columns.
On the contrary, TAV is able to achieve relatively better temporal consistency, thanks to its temporal modeling and one-shot tuning. However, the pose and the motion in the edited video are not faithful to the original video. 
Our method can take the best of both worlds, and successfully turn the horse into a dog while maintaining both temporal consistency and faithfulness to the input video.

\figuser
\paragraph{Quantitative Comparison.}
\textit{(a) User Preference}.  Following~\cite{wu2022tune,singer2022make}, we conduct user preference studies on 32 videos to compare our method with TAV and PnP, respectively. The users are asked to select editing better results based on (1) quality, (2) text-to-video alignment, and (3) fidelity to the original video. The results are shown in the first two columns in Fig.~\ref{fig:user}. 
As can be seen, we achieve higher scores in all three indicators when compared to PnP and TAV. 
\textit{(b) CLIP score}~\cite{hessel2021clipscore,esser2023structure} measures text-to-video alignment by computing the similarity between the target prompt and the edited videos. As can be seen in the third column in Fig.~\ref{fig:user}, our method obtains an on-par CLIP score to that of TAV  finetuned on the input video. We should note that the automatic evaluation metrics like CLIP score are not always consistent with human perception~\cite{molad2023dreamix}, and should only be used as an imperfect score for reference. 
\textit{(c) Frame consistency}~\cite{esser2023structure} measures the consistency of the generated video. We follow~\cite{esser2023structure} to calculate the cosine feature similarity of consecutive frame pairs, based on their CLIP embedding, and take the average result as the score. As can be seen in the third column of Fig.~\ref{fig:user}, our method obtains the best score in these three methods, which demonstrates the superiority of our method in maintaining temporal consistency.

\subsection{Applications on Customized Video Editing}

\begin{figure*}[t]
\centering
\includegraphics[width=\textwidth]{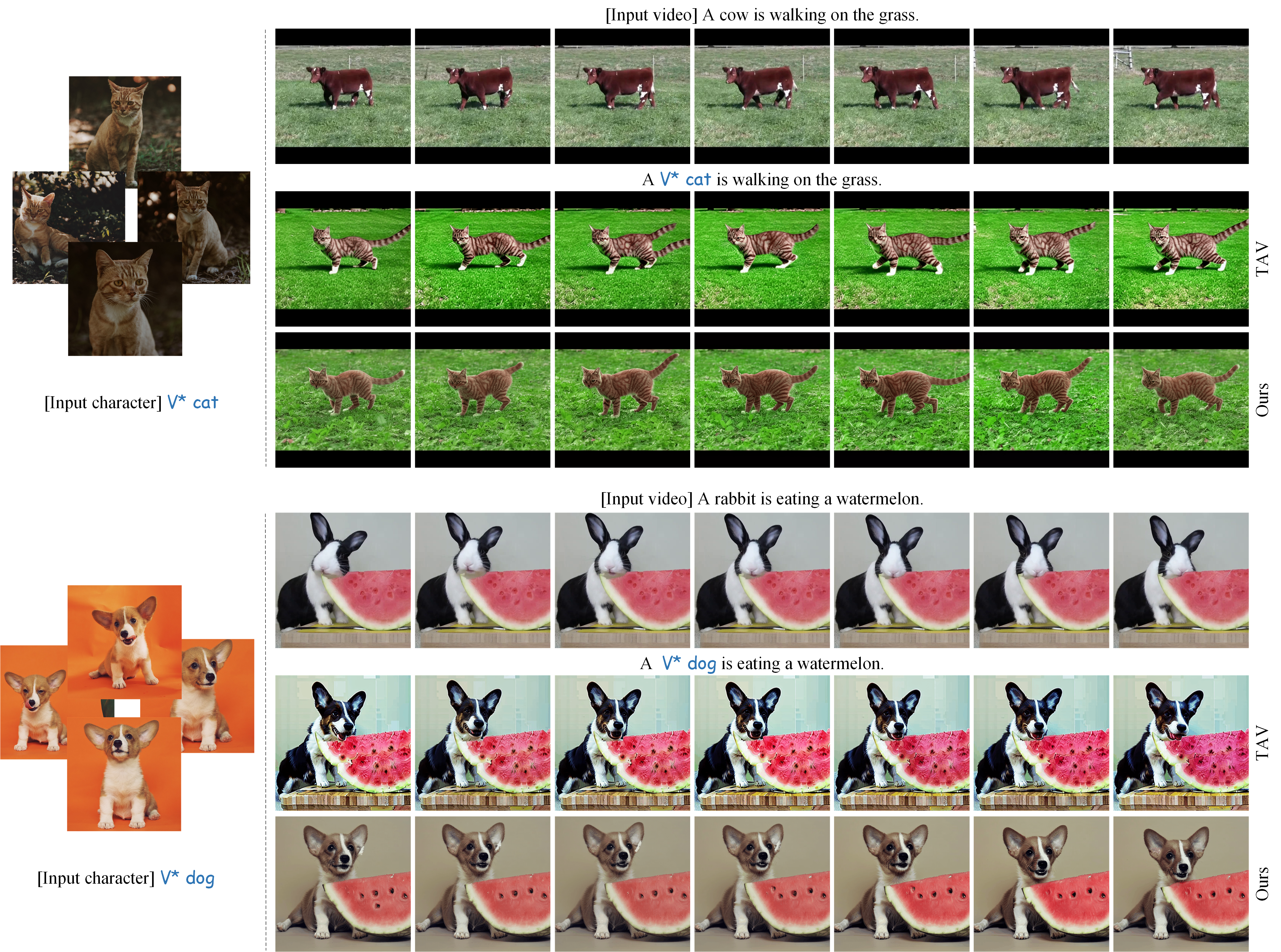}
\caption{Applications on customized video editing. The images of customized characters are shown on the left, while the input and edited videos and shown on the right.}
\label{fig:cust}
\end{figure*}

In this section, we extend vid2vid-zero to the task of customized video editing. Customized video editing aims to replace objects in a video with objects of specific identity that are of interest to the user. Benefiting from our use of existing image diffusion models for video editing, we can directly realize customized video editing using customized image generation models. Specifically, we use DreamBooth~\cite{ruiz2022dreambooth} to fine-tune the input customized character images and use the resulting model for video customization editing. For comparison, we extend the same approach to TAV~\cite{wu2022tune}, \textit{i.e.}, we use the same customized image generation model to fine-tune the single input video and perform customized video editing with the trained model.

The results are shown in Fig.~\ref{fig:cust}. It can be seen that vid2vid-zero can replace the foreground object with a customized object of interest, \textit{e.g.}, replacing a cow with an orange cat shown on the left side, or replacing a rabbit with a dog shown on the left side. While TAV is also able to preserve the identity of the customized objects to some extent, it faces several artifacts, such as the cat's tail repeating itself and the dog's color changing (from yellow to black). We speculate that this is because TAV fine-tunes the customized image model, resulting in the loss of the identity information. In contrast, our vid2vid-zero avoids the above problems and achieves better customized editing results.

\section{Conclusion and Discussion}
In this paper, we propose vid2vid-zero, a method that achieves zero-shot video editing using off-the-shelf image diffusion models, without training on any video data. 
We leverage the dynamic nature of the attention mechanism to enable effective test-time temporal modeling.
Experiments show that vid2vid-zero preserves the creativity of the image diffusion model and can obtain high-quality, text-aligned, and faithful video editing results. With the rapid development of generative models, we hope to bring novel insights into the applications of these models.

\textbf{Limitations.}
Since our method directly uses the pre-trained weights of the image diffusion model, it may inherit the limitations of the off-the-shelf image generation model. For example, our method lacks temporal and motion priors, because the off-the-shelf image diffusion model is not trained on any video data. As a result, it cannot be directly used to edit actions in videos, as reflected by our inability to effectively edit verbs in the source prompts. 



\bibliographystyle{IEEEtran}
\bibliography{ref}


\clearpage

\vfill
\end{document}